\documentclass[manuscript,screen]{acmart}
\AtBeginDocument{%
  \providecommand\BibTeX{{%
    \normalfont B\kern-0.5em{\scshape i\kern-0.25em b}\kern-0.8em\TeX}}}

\setcopyright{acmlicensed}
\copyrightyear{2024}
\acmYear{2024}
\acmDOI{XXXXXXX.XXXXXXX}

\acmConference[Conference undisclosed for anonymity]{Conference undisclosed for anonymity}{}{}
\acmBooktitle{Conference undisclosed for anonymity} 
\usepackage{pifont,multirow}
\usepackage{enumitem}
\usepackage[compact]{titlesec}
\titlespacing{\section}{0pt}{5pt}{5pt}
\titlespacing{\subsection}{0pt}{5pt}{5pt}
\titlespacing{\subsubsection}{0pt}{5pt}{5pt}
\begin{document}

%%
%% The "title" command has an optional parameter,
%% allowing the author to define a "short title" to be used in page headers.
\title[How VADER is your AI?]{How VADER is your AI? Towards a definition of artificial intelligence systems appropriate for regulation}

%%
%% The "author" command and its associated commands are used to define
%% the authors and their affiliations.
%% Of note is the shared affiliation of the first two authors, and the
%% "authornote" and "authornotemark" commands
%% used to denote shared contribution to the research.
\author{Leonardo C. T. Bezerra}
\email{leonardo.bezerra@stir.ac.uk}
\orcid{0000-0003-4654-2553}
\affiliation{%
  \institution{University of Stirling}
  \streetaddress{YYYYY}
  \city{Stirling}
  \state{YYYY}
  \country{United Kingdom}
  \postcode{00000-000}
}

\author{Alexander E. I. Brownlee}
\email{alexander.brownlee@stir.ac.uk}
\orcid{0000-0003-2892-5059}
\affiliation{%
  \institution{University of Stirling}
  \streetaddress{YYYYY}
  \city{Stirling}
  \state{YYYY}
  \country{United Kingdom}
  \postcode{00000-000}
}

\author{Luana Ferraz Alvarenga}
\email{lferraza@gmail.com}
\orcid{YYYY}
\affiliation{%
  \institution{ Federal University of Rio Grande do Norte}
  \streetaddress{YYYYY}
  \city{Natal}
  \state{YYYY}
  \country{Brazil}
  \postcode{00000-000}
}

\author{Renan Cipriano Moioli}
\email{renan.moioli@imd.ufrn.br}
\orcid{0000-0001-6036-8358}
\affiliation{%
  \institution{ Federal University of Rio Grande do Norte}
  \streetaddress{YYYYY}
  \city{Natal}
  \state{YYYY}
  \country{Brazil}
  \postcode{00000-000}
}

\author{Thais Vasconcelos Batista}
\email{thais@dimap.ufrn.br}
\orcid{YYYY}
\affiliation{%
  \institution{ Federal University of Rio Grande do Norte}
  \streetaddress{YYYYY}
  \city{Natal}
  \state{YYYY}
  \country{Brazil}
  \postcode{00000-000}
}

%%
%% By default, the full list of authors will be used in the page
%% headers. Often, this list is too long, and will overlap
%% other information printed in the page headers. This command allows
%% the author to define a more concise list
%% of authors' names for this purpose.
\renewcommand{\shortauthors}{Bezerra et al.}

%%
%% The abstract is a short summary of the work to be presented in the
%% article.
\begin{abstract}
Artificial intelligence (AI) has driven many information and communication technology (ICT) breakthroughs. Nonetheless, the scope of ICT systems has expanded far beyond AI since the Turing test proposal. Critically, recent AI regulation proposals adopt AI definitions affecting ICT techniques, approaches, and systems that are not AI. In some cases, even works from mathematics, statistics, and engineering would be affected. Worryingly, AI misdefinitions are observed from Western societies to the Global South. In this paper, we propose a framework to score how \textit{validated as appropriately-defined for regulation} (VADER) an AI definition is. Our online, publicly-available VADER framework scores the coverage of premises that should underlie AI definitions for regulation, which aim to (i) reproduce principles observed in other successful technology regulations, and (ii) include all AI techniques and approaches while excluding non-AI works. Regarding the latter, our score is based on a dataset of representative AI, non-AI ICT, and non-ICT examples. We demonstrate our contribution by reviewing the AI regulation proposals of key players, namely the United States, United Kingdom, European Union, and Brazil. Importantly, none of the proposals assessed achieve the appropriateness score, ranging from a revision need to a concrete risk to ICT systems and works from other fields.
\end{abstract}

%%
%% The code below is generated by the tool at http://dl.acm.org/ccs.cfm.
%% Please copy and paste the code instead of the example below.
%%


%%
%% Keywords. The author(s) should pick words that accurately describe
%% the work being presented. Separate the keywords with commas.

\received{22 January 2024}
\received[revised]{--}
\received[accepted]{--}

%%
%% This command processes the author and affiliation and title
%% information and builds the first part of the formatted document.
\maketitle

\section{Introduction}
\label{Sec:Introduction}
Information and communication technologies~(ICT) have been central in shaping our society for nearly a century now, with novel breakthrough technologies becoming ubiquitous and disrupting social and commercial relations periodically~\cite{Inaba2017}. The rising importance of ICT in society is further evidenced by its classification by the \textit{International Standard Classification of Education}~(ISCED-F) as one of ten broad fields of study~\cite{unesco2013isced}, achieving the same level of importance of well-established fields such as education and engineering. The rapid evolution of ICT over time is illustrated by the continuous update of ICT curricula, with disciplines often becoming mature enough to comprise fields of study themselves. The \textit{Association for Computing Machinery}~(ACM) currently identifies six major deeply interconnected ICT curricula worldwide~\cite{acm2020cc}, namely computer engineering, computer science, cybersecurity, information systems, information technology, and software engineering. In part, the constantly evolving landscape in ICT is made possible by the little regulation existing in the field, especially when compared to other traditional fields.

Since the early days of ICT research, the pursuit of \textit{artificial intelligence}~(AI) has been set as a possibility and/or a goal in the field. Originally, AI research was heavily defined by the \textit{Turing test}, proposed as a practical exercise to assess whether machines could think~\cite{Turing1950}. Indeed, the early literature in ICT is deeply intertwined with the early literature in AI, and the potential accredited to AI drove a number of ICT breakthroughs~\cite{russell2020}. Over the years, ICT fields matured significantly to become independent even if deeply interconnected, as previously discussed. In turn, AI is currently classified either as a discipline of computer science~\cite{acm2020cc} or as a field of study itself~\cite{unesco2013isced}, both classifications acknowledging the clear distinction between AI and non-AI ICT.

The recent breakthroughs in AI have stirred the need for a structured regulation of AI systems, in the wake of successful data privacy regulations~\cite{GDPR2016a} and the speed with which AI systems started being developed and deployed since the advent of \textit{deep learning}~\cite{LeCun2015}. Importantly, the disruptive potential demonstrated by \textit{generative AI} since the introduction of ChatGPT~\cite{openai2022chatgpt} and similar systems have greatly accelerated regulation proposals from key AI players. In Western societies, proposals have already been introduced by the United States~\cite{us2019aaa,us2022aaa,us2023aaa,us2023eo,us2023airia,us2024ccpaw}, the United Kingdom~\cite{uk2023bill}, and the European Union~\cite{eu2021aiact}, for instance. Expectedly, countries in the Global South have followed this pattern, and Brazil has already preliminarily approved its own regulation on AI~\cite{br2020pl21,br2023pl2338}. In common, those proposals attempt to improve AI fairness, accountability, and transparency through artifacts such as model cards and processes like evaluation and monitoring. Importantly, some proposals establish the need for human oversight, an approach that is being highly debated given its potential impact on the scalability benefits provided by the AI systems. 

However welcome, the existing AI regulation proposals have not adopted a unified definition of AI systems. Worryingly, some of the definitions in use are so broad that they fail to distinguish between AI and non-AI ICT systems, and even include works from other fields such as mathematics, statistics, and engineering. In part, this is a consequence of the plurality of AI definitions observed in the literature~\cite{russell2020}, stemming from (i)~the~interdisciplinary nature and (ii)~the~diversity in techniques and approaches in AI. Concerning the former, AI research is deeply intertwined with other sciences such as cognitive sciences. As such, fundamental concepts such as intelligence and consciousness are difficult to objectively define. Regarding the latter, AI techniques are often subject to the \textit{no free lunch theorems}~\cite{WolMac1995search,Wol1996learning,WolMac1997opt,Wol2001sl}, which imply that AI algorithms will perform differently as a function of the problem tackled. Given the diversity of problems that AI research addresses, it is only natural that an even more diverse set of techniques and approaches be proposed over the years. Importantly, we observe the increased interest in AI magnifying this problem, with novel techniques and approaches being proposed at an alarming speed~\cite{AraCamCam2022}.

In this paper, we propose a framework to score how \textit{validated as appropriately-defined for  regulation}~(VADER) an AI definition is. Our approach scores the coverage of premises that should underlie AI definitions for regulation, labeling definitions as (i)~\textit{appropriate for regulation}; (ii)~\textit{revision needed}, or; (iii)~\textit{concrete risk}. Regarding premises, a first subset aims to reproduce principles observed in successful worldwide data privacy regulation, namely (i)~\textit{technology neutrality} and (ii)~\textit{technical unambiguity}. The second subset of premises aims to correctly delimit the scope of the regulation, namely that the definition (iii)~be~defined in \textit{objective} terms; (iv)~exclude works from fields \textit{other than ICT}; (v)~exclude ICT techniques, approaches, and systems that are \textit{not connected to AI}, and; (vi)~include \textit{all} AI techniques and approaches. To assess the scope of a given definition, we propose a dataset of representative examples of AI, non-AI ICT, and non-ICT techniques, approaches, systems, and/or works. Based on existing taxonomies and policy recommendations, this dataset helps regulators identify eventual scope issues to work on. 

To demonstrate the benefits of our proposed framework, we critically review the AI definitions adopted by existing regulation proposals from key AI players in Western societies and in the Global South. Regarding the former, we assess proposals from the US~\cite{us2023aaa}, UK~\cite{uk2023bill}, and EU~\cite{eu2021aiact}. Concerning the latter, we assess a proposal from Brazil~\cite{br2023pl2338}, which we select for its relevance both in the Global South and in Latin America and its successful history of technological regulation. Worryingly, none of the proposals assessed achieve the appropriate-for-regulation score. The issues observed mostly concentrate on scope, and range from (i) \textit{revision needed}, when some AI techniques or approaches are not included in the definition adopted, to; (ii) \textit{critical risk}, when works not connected to AI fit the broad definitions adopted. Unfortunately, we observe the latter scenario in all countries assessed. For instance, the UK AI Act~\cite{uk2023bill} defines AI systems as a function of ``\textit{cognitive abilities}'', leaving significant room for subjectivity. In turn, the only type of non-AI ICT systems that the US Algorithmic Accountability Act~\cite{us2023aaa} excludes is what they label ``\textit{passive computing infrastructure}''. Importantly, we discuss how recent revisions of the EU AI Act and Brazilian PL have reduced the risk of overregulation.%~\cite{GilMurWri2021}. 

The contributions of our work can be summarized as follows:
\begin{enumerate}
    \item A validation framework to assess the appropriateness for regulation of AI definitions, to be made publicly available online for governments, industry, academia, and civil society upon acceptance.
    %\footnote{The link to the online framework page is omitted during review not to compromise anonymity.}
    \item The identification of premises from technical and policy literature to underlie AI definitions for regulation.
    \item A dataset of representative examples of (i) AI techniques and approaches; (ii) non-AI ICT techniques, approaches, and systems, and; (iii) non-ICT works, which help identify scope issues with proposed AI definitions. 
    \item A critical review of the current worldwide AI regulation proposals from key AI players w.r.t their AI definitions, ranging from Western societies to the Global South, with recommended actions to be taken.  
\end{enumerate}

The remainder of this work is organized as follows. Section~\ref{Sec:Background} provides background in ICT, AI, and successful technology regulation. In Section~\ref{Sec:Towards}, we propose our validation framework, detailing the dataset of representative examples, the set of premises identified in the literature, and the scoring we draw from these two factors. Later, Section~\ref{Sec:critical} demonstrates the benefits of our proposal by critically reviewing worldwide AI regulation proposals. Finally, we conclude and discuss future work in Section~\ref{Sec:Conclusion}.
\section{Background}
\label{Sec:Background}

Artificial intelligence~(AI) regulation is a topic that requires background into multiple fields of study, yet engages a plural audience that will seldom be familiar with all the concepts required for its understanding. In this section, we provide a broad overview of the main concepts that connect to AI regulation, using existing policy recommendations and taxonomies to introduce information and communication technologies, AI, and technology regulation. 

\subsection{Information and Communication Technologies~(ICT)}

ICT is a fast-evolving field of study, as evidenced by the multiple names the field has been known for over the decades. To provide an overview of the field as it currently stands, we initially refer to existing policy (recommendations) from the \textit{Association of Computing Machinery}~(ACM) and the \textit{United Nations Educational, Scientific and Cultural Organization} (UNESCO), respectively the \textit{ACM Computing Curricula 2020}~\cite{acm2020cc} and the \textit{International Standard Classification of Education}~(ISCED-F)~\cite{unesco2013isced}. In particular, we list below the ICT detailed fields of study taken from the Brazilian revision of ISCED-F~\cite{inep2019cine}, adopted not only for statistical analysis but also as a policy for higher education regulation. Importantly for this work, the revision combines the works of ACM and UNESCO and resulted from a collaboration between academia, government, and the Brazilian Computing Society.

\begin{description}[topsep=5pt,leftmargin=0px,style=unboxed]
    \item [ICT infrastructure and management.] Concerns the management of ICT and required infrastructure. Relevant examples include databases, cybersecurity, and networks.
    \item [Information systems management and development.] Regards understanding, analyzing, and solving social and organizational problems through the creation, management, and evaluation of information systems. Relevant examples include information systems, information security, and web development.

    \item [Software production.] Concerns the management, engineering, and production of software. Relevant examples include software engineering and game development.
    \item [Computer science.] Refers to the theoretical, scientific, and technological development of computing. Relevant examples include computer science and artificial intelligence.
    \item [Development of systems that integrate hardware and software.] Regards the design and implementation of hardware components and systems that integrate hardware and software in domains such as automation, control, and robotics. Relevant examples include computer engineering, embedded systems, and internet-of-things. 
\end{description}

In terms of systems, the above definitions evidence how plural their nature can be. In detail, top-most definitions regard ICT systems that are closer to business and society, and bottom-most definitions refer to ICT systems that are closer to foundation or physical solutions. In this context, AI systems can fit the remit of computer science when they are conceived as foundation models, but the remit of information systems once applied to a social or commercial context. In addition, AI systems may be developed embedded in or in connection to a physical system, require appropriate infrastructure, and adopt software engineering methodologies for operations and accountability. Appendix~\ref{Sec:SoS} further details systems, their taxonomies and their integration as systems-of-systems~(SoS), i.e. their real-world organization.

\subsection{Artificial intelligence (AI)}

As discussed above, AI is a field of study deeply connected to computer science~(CS). Notwithstanding, AI is classified by ACM and UNESCO differently: ACM views AI as a CS discipline, whereas UNESCO sees AI as a transversal field of study. From a high-level perspective, the UNESCO ISCED-F classification stems from the interdisciplinary nature of AI that leads to a plurality of AI definitions, as we next discuss. For brevity, Appendix~\ref{Sec:AI} discusses AI from the perspective of the \textit{ACM Computer Science 2023 Curriculum}~\cite{acm2023cs}, particularly the diversity in AI techniques and approaches.
%
% \subsubsection{Definitions}
%
AI academic definitions can be grouped as a function of whether they consider (i) thought or behavior and (ii) human or rational performance \cite{russell2020}. The combination of these factors lead to the following four perspectives:

\begin{description}[style=unboxed,leftmargin=0px,topsep=5pt]
    \item[Human behavior.] The original approach to AI is tightly related to this perspective, illustrated by the \textit{Turing Test}~\cite{Turing1950}. Many AI techniques and approaches stemmed from the goals set by this perspective, most notably search, knowledge representation, automated reasoning, machine learning, perception, and robotics.
    \item[Human thought.] This AI perspective connects AI and cognitive sciences to model and replicate human reasoning. From this perspective, AI is the computational tool to achieve an interdisciplinary goal of understanding intelligence, consciousness, and other subjective concepts that computational models can help address.
    \item[Rational thought.] Also known as logicist AI, this perspective focuses on formally defining problems in logical notation and automating their resolution. Importantly, this perspective may provide correctness when feasible.
    \item[Rational behavior.] This perspective is based on the concept of agents, i.e., computer systems that \textit{"operate autonomously, perceive their environment, persist over a period of time, adapt to change, and create and pursue goals"}. The objective of a rational agent is to achieve either the best or a high-quality outcome for the problem it addresses. 
\end{description}

In the context of policy, the thought-oriented AI perspectives present less pressing concerns. In fact, the rational thought perspective can help AI system explainability and therefore fairness, accountability, and transparency. Regarding behavior-oriented perspectives, the human behavior perspective is interested in replicating human performance, which in general can be embedded with bias. In turn, the rational behavior perspective drives AI development from the perspective of objective measures. In practice, both behavior-oriented perspectives are necessary for regulation, as existing fairness measures present conflicts in representing, if at all, human definitions of fairness \cite{Sampat2019}. We refer to Appendix~\ref{Sec:Definitions} for further details on how existing AI policy recommendations from diverse worldwide organizations relate to the definition perspectives discussed in this section, but highlight that overall (i)~definitions follow the rational behavior perspective and (ii)~definitions from the same organization may follow different perspectives.

\subsection{Tech regulation}

In the domain of ICT regulations, substantial global variations exist, primarily due to diverse legal, cultural, and social considerations that motivate individual countries to establish regulatory frameworks. Factors such as national security, political governance, cultural and social advancement, technological progress, economic considerations, and geopolitical dynamics can exert significant influence on the progression of regulatory initiatives. Moreover, the dynamic nature of the ICT domain, characterized by continuous technological evolution, poses inherent challenges to the regulatory development process. The pace of technological advancements requires a flexible and adaptive approach to regulation. Typically, ICT regulations cover a diverse spectrum of dimensions, including telecommunications, internet services, data protection, cybersecurity, intellectual property, and other relevant areas.

A relevant niche of successful ICT regulation addresses the use of personal data. As data became an essential tool for the development of diverse business models, companies across sectors started to collect significant data amounts. Consequently, the need to regulate ownership and use of personal data did not go unnoticed by legislators worldwide. In this review, we detail three personal data laws that have been implemented by the countries we will later assess: (i)~the~EU~General Data Protection Regulation~(GDPR~\cite{GDPR2016a});%
\footnote{The EU GDPR has its UK law incorporation in the UK GDPR, with ``\textit{little change to the core data protection principles, rights and obligations}''~\cite{uk2021ico}.} 
(ii)~the~California Consumer Privacy Act~(CCPA~\cite{us2018ccpa}), and; (iii)~the~Brazilian General Data Protection Law~(LGPD~\cite{br2018lgpd}). Given their similarities, we focus here on GDPR, and refer to Appendix~\ref{Sec:Regulation} for the latter two. 

GDPR serves as a paradigm of ICT regulation for several reasons, primarily due to its global impact and influence. Despite originating as an EU regulation, it has an extraterritorial reach, subjecting any organization that processes the personal data of EU residents, regardless of its location. The implementation of GDPR has influenced a global discussion on privacy and data protection, inspiring legislative efforts in other countries and regions, some of which have adopted similar regulations. Importantly, GDPR is the first legal text to articulate the implementation of \textit{data protection by design} as a principle and an obligation, incorporating data security into technology from its inception. This approach safeguards the rights of natural persons, positioning them as managers of their own personal data. 
Another relevant aspect of GDPR is the recognition of the importance and the influence of technology in different matters, especially to social and economic progress, emphasizing the need of the responsible development of technology. The technological aspect of data protection cannot be ignored, and to legislate such a dynamic matter effectively the GDPR emphasizes the importance of technology neutrality in its recitals. The adoption of the technology-neutral legislation shows the flexibility of the code, enabling the regulation to remain pertinent even with the development of ICT. 

\smallskip
As discussed in this section, ICT is a broad field of study with deeply interconnected curricula that are constantly evolving. In this context, AI is either classified as a discipline within computer science or as a transversal field of study due to its interdisciplinary nature and the diversity in AI techniques and approaches. Altogether, these factors pose a significant challenge for AI regulation. Nonetheless, broad knowledge into ICT, AI, and successful worldwide tech regulation examples provide insights for how AI regulation proposals may succeed. In the next section, we structure those insights as a framework to validate the appropriateness for regulation of AI definitions.

\section{Towards appropriate-for-regulation AI definitions}
\label{Sec:Towards}

AI regulation has become one of the most important topics for governments, industry, academia, and civil society due to the expected disruptions that recent AI technologies have introduced. In particular, the developments in deep learning~\cite{LeCun2015} in the 2010’s and multimodal large language models such as ChatGTP~\cite{openai2022chatgpt} in the 2020’s demonstrated how fast novel AI technologies and systems can have a striking social and commercial impact. Nonetheless, the challenges previously discussed for AI regulation result in proposals that are still being matured by governments, requiring the effective participation of industry, academia, and civil society. 

In order to assist AI regulation, we propose a framework to validate the appropriateness for regulation~(VADER) of an AI definition.  Concretely, our approach scores the coverage of premises that should underlie AI definitions for regulation, labeling definitions as (i)~appropriate for regulation; (ii)~needing revision, or; (iii)~posing concrete risk of overregulation. Premises aim to reproduce principles observed in successful worldwide data privacy regulation and to correctly delimit the scope of the regulation. To assess scope, we propose a dataset of representative examples of AI, non-AI ICT, and non-ICT techniques, approaches, systems, and/or works. Based on existing taxonomies and policy recommendations, this dataset helps identify eventual scope issues so regulators know where to concentrate their revisions on. To be didactic, in this section (i) we address the constituent elements of our framework in reverse order, starting from the representative dataset and ending in the scoring outcomes, and; (ii) use the OECD AI definitions as a case study~\cite{oecd2019ai,oecd2023ai}, which will later underlie our assessment of worldwide AI regulation.

\subsection{Representative example dataset}

%TABLE%
\begin{table*}
  \caption{Techniques and approaches from ICT disciplines other than AI or non-ICT fields that AI rational agents may employ.}
  \label{tab:techniques}
  \scalebox{0.8}{
  \begin{tabular}{p{2.8in}p{2in}p{1.7in}}
    \toprule
    \textbf{Property} & \textbf{Techniques and approaches} & \textbf{Associated discipline or field}\\
    \midrule
    Problem solving and search & Uninformed search & Algorithms \& complexity (ICT)\\
    Knowledge representation, reasoning and planning & Propositional and fuzzy logic&Philosophy\\
    Learning&Distribution-driven analysis&Statistics\\
    Natural language&Parsing&(Computational) Linguistics\\
    Perception&Signal processing&Mathematics and engineering\\
    Robotics&Control algorithms&Engineering\\
    \bottomrule
\end{tabular}
}
\end{table*}

As discussed in the background, AI rational agents may employ techniques and approaches from ICT disciplines other than AI or even non-ICT fields when addressing problems. These are summarized in Table \ref{tab:techniques}, which shows a range of non-ICT fields from philosophy and linguistics~(arts and humanities) to statistics, mathematics, and engineering~(STEM). As we will later discuss, characterizing a system as AI solely because of the adoption of those techniques would risk compromising works from fields with an extensive literature independent of AI. To address this issue, we propose a dataset of representative examples comprising (i) AI techniques and approaches; (ii) non-AI ICT techniques, approaches, and systems, and; (iii) non-ICT works, which concretely help identify scope issues with proposed AI definitions for regulators to address. The dataset is detailed in Appendix \ref{Sec:Dataset}, and the policy recommendations that underlie the examples listed for each category are further discussed below.

\begin{description}[style=unboxed,leftmargin=0px,topsep=5pt]
\item[AI techniques and approaches.] The examples we select reflect the properties discussed in the AI textbook work of~\citet{russell2020}, which are also representative of the \textit{ACM Computer Science Curriculum~2023} for the AI discipline~\cite{acm2023cs}. Importantly, the examples in this category are not \textit{required} to qualify a system as AI, but are \textit{sufficient} to such qualification as only AI systems adopt them.

\item[Non-AI techniques, approaches, and systems.] This group of examples reflects multiple taxonomies. For techniques and approaches, we adopted the (i)~\textit{ACM~Computing Curricula~2020}~\cite{acm2020cc} and (ii)~\textit{ACM~ Computer Science Curriculum~2023}~\cite{acm2023cs}. For systems, we focus on the ``\textit{Large-capacity information analysis}" of the ``J tag'' ICT systems taxonomy~\cite{Inaba2017}, discussed in Appendix~\ref{Sec:SoS}, which we further enrich with example from ACM~curricula disciplines. By themselves, none of these examples qualifies as AI systems, \textit{unless} they adopt the sufficient AI techniques and approaches given in the previous category. As such, the large number of examples in this category evidences how important it is to differentiate between AI and non-AI ICT.

\item[Non-ICT works.] The examples in this category reflect the discussion depicted in Table \ref{tab:techniques}, referencing corresponding literature. Importantly, we also include examples of processes and products from engineering fields like electronic, electrical, and mechanical engineering that could erroneously be affected by regulation adopting a broad AI definition.
\end{description}

\subsection{Premises}
As previously discussed, the proper regulation of technology requires premises to be respected by their proponents. In particular, the definition of AI systems is one of the most critical aspects concerning AI regulation, given the plurality of AI definitions available in the academic literature and in existing policy recommendations. Below, we discuss the premises that an AI definition appropriate for regulation should present. Premises are grouped by their purpose, namely (i)~reproducing principles observed in successful worldwide data privacy regulation~(Premises~1--2), and (ii)~correctly delimiting the scope of the regulation~(Premises~3--6). As a concrete case study to ground our discussion, we consider the OECD 2019 and 2023 definitions~\cite{oecd2019ai,oecd2023ai}, respectively given below. 
% For brevity, all premises are listed below, but justifications and discussions for premises that are not violated by OECD definitions are given in Appendix~\ref{Sec:Premises}. 

\newtheorem{oecd}{OECD Definition}
\setcounter{oecd}{2018}
\begin{oecd}
% \begin{quote}
An AI system is a machine-based system that can, for a given set of human-defined objectives, make predictions, recommendations, or decisions influencing real or virtual environments. AI systems are designed to operate with varying levels of autonomy.    
\end{oecd}
% \end{quote}

\setcounter{oecd}{2022}
\begin{oecd}
% \begin{quote}
An AI system is a machine-based system that, for explicit or implicit objectives, infers, from the input it receives, how to generate outputs such as predictions, content, recommendations, or decisions that can influence physical or virtual environments. Different AI systems vary in their levels of autonomy and adaptiveness after deployment.
\end{oecd}
% \end{quote}

\newtheorem{premise}{Premise}

\setcounter{premise}{0}
% \begin{premise}
% \textbf{Technology agnosticity.} Definitions should not include technologies which can become obsolete in time.
% \end{premise}
% \noindent \textit{Justification}. The rapid evolution of ICT incurs in the proposal of novel technologies each year. As such, an AI regulation that is not agnostic will need constant revision, which may altogether compromise its effectiveness due to bureaucracy.

% \noindent \textit{Discussion}. The original OECD 2019 definition is high-level enough to be agnostic to rapid technology development. The updates introduced in the OECD 2023 definition did not compromise its technology agnosticity.

\begin{premise}
\textbf{Technology neutrality.} Definitions should not be in favor or in detriment of specific sectors of the market, nor include technologies which can become obsolete in time~\cite{GDPR2016a}.
\end{premise}         

\noindent \textit{Justification}. Technology regulation can affect the market when some technologies are favored in detriment of others. Collaboration with a wide range of industry sectors is required to ensure neutrality in the AI definition agreed upon. Furthermore, the rapid evolution of ICT incurs in the proposal of novel technologies each year. As such, an AI regulation that is not agnostic will need constant revision, which may altogether compromise its effectiveness due to bureaucracy.

\noindent \textit{Discussion}. The original OECD 2019 definition is high-level enough to be agnostic to rapid technology development, and do not affect the market in any particular way. The updates introduced in the OECD 2023 definition did not compromise its technology neutrality.

\begin{premise}
\textbf{Technical unambiguity.} Definitions should employ only technical terms that are unambiguous.
\end{premise}

\noindent \textit{Justification}. Technology regulation poses a challenge for regulators from diverse backgrounds due to the need of precise and accurate technical terms. If ambiguous or imprecise terms are employed, the scope and implications of the regulation can be significantly affected.

\noindent \textit{Discussion}. The original OECD 2019 definition is precise enough not to lead to confusion about the terms it employs. The updates introduced in the OECD 2023 definition did not compromise its technical unambiguity.

\begin{premise}
\textbf{Objectivity.} Definitions should adopt only objective terms, i.e. the rational behavior perspective~\cite{russell2020}. 
\end{premise}

\noindent \textit{Justification}. The rationale behind this premise is to avoid the inclusion of subjective terms. Adopting the rational behavior perspective avoids stirring a discussion on complex subjects such as intelligence, consciousness, and cognition.\sloppy

\noindent \textit{Discussion}. The original 2019 OECD definition is objectively defined in terms of the rational behavior perspective. The updates introduced in the OECD 2023 definition did not compromise its objectivity.

% \begin{premise}
% \textbf{Technology agnosticity.} Definitions should not include technologies which can become obsolete in time.
% \end{premise}
% \begin{premise}
% \textbf{Technology neutrality.} Definitions should not be in favor or in detriment of specific sectors of the market.
% \end{premise}         
% \begin{premise}
% \textbf{Technical unambiguity.} Definitions should employ only technical terms that are unambiguous.
% \end{premise}
% \begin{premise}
% \textbf{Objectivity.} Definitions should adopt only objective terms, i.e. the rational behavior perspective~\cite{russell2020}. 
% \end{premise}
\begin{premise}
\textbf{Exclusion of non-ICT works.} Works such as techniques, approaches, products, and processes from non-ICT fields should be excluded from the scope of the definition.  
\end{premise}

\noindent \textit{Justification}. Two different situations motivate this premise. The first concerns definitions where a (non-)exhaustive list of techniques and approaches qualifies an AI system. Often, such a list includes the works given in Table~\ref{tab:techniques}. Relevant examples are the terms (i) \textit{optimization}, which incorrectly includes all of \textit{mathematical} optimization~\cite{GilMurWri2021} instead of only \textit{heuristic} optimization~\cite{Hoos2005}, and; (ii) \textit{statistics} or \textit{statistical approaches}, which incorrectly include the statistics literature that assumes that the data are generated by a given stochastic data model instead of only the statistics literature that uses algorithmic models and treats the data mechanism as unknown~\cite{breiman2001}. The second situation concerns definitions where no list of techniques and approaches is adopted. In this case, the definition often becomes so general that it may include works from other non-ICT fields. The most straightforward examples come from engineering, where works from fields like electrical, electronic, and mechanical engineering may fit definitions incorrectly. 

\noindent \textit{Discussion}. The original 2019 OECD definition does not list techniques and approaches as an AI-system qualifier. Furthermore, the expression ``\textit{machine-based system that can (...) make (...) decisions influencing real (...) environments}'' is so broad that many engineering devices would fit this description. Examples from our validation dataset range from simple electronic devices like a thermostat to classical mechanical engineering systems, such as the steam engine governor. The updates introduced in the OECD 2023 definitions did not make the novel definition compliant with this premise, as it includes the equivalent expression ``\textit{system that (...) infers (...) how to generate outputs such as (...) decisions (...) influencing physical (...) environments}''.

\begin{premise}
\textbf{Exclusion of non-AI ICT techniques, approaches, and systems.} Techniques and approaches from ICT disciplines other than AI should be excluded, as well as systems that do not employ AI techniques and approaches.
\end{premise}

\noindent \textit{Justification}. The two different scenarios that motivated the previous premise also motivate this premise. In detail, (non-)exhaustive list of techniques and approaches that qualify an AI system often adopt terms that are too broad, including techniques and approaches that originate in an ICT discipline other than AI. A relevant example is search, which incorrectly includes uninformed techniques that originate in the \textit{algorithms} and \textit{complexity} ICT discipline \cite{Cormen2022}. In turn, when no list is provided, broad definitions may erroneously include ICT systems that in no way make use of AI techniques and approaches. One relevant example is to refer to AI systems as systems that are employed to assist in decision-making, which not necessarily are AI systems.

\noindent \textit{Discussion}. As previously mentioned, the original 2019 OECD definition does not include an AI-system qualifier list of techniques and approaches. In addition, the expression ``\textit{system that can (...) make (...) recommendations (...) influencing real (...) environments}'' is broad enough that many ICT systems employed to assist decision-making would qualify. Examples from our validation dataset include enterprise resource planning systems, as well as information systems dashboards. The updates introduced in the OECD 2023 definitions did not make the novel definition compliant with this premise, as it includes the equivalent expression ``\textit{system that (...) infers (...) how to generate outputs such as (...) recommendations (...) influencing physical (...) environments}''.

\begin{premise}
\textbf{Inclusion of all AI techniques and approaches.} All techniques and approaches that can qualify a system as AI should be included.
\end{premise}

\noindent \textit{Justification}. The two different scenarios that motivated the previous premises also motivate this premise.  In detail, (non-)exhaustive list of techniques and approaches that qualify an AI system sometimes oversee AI techniques and approaches. One example is overly focusing on learning and missing out on other aspects of AI, such as search and reasoning. In turn, when no list is provided, broad definitions can very easily defined AI systems in a way that only accounts for a subset of techniques and approaches.

\noindent \textit{Discussion}. As mentioned, the 2019 OECD definition does not include an AI-system qualifier list of techniques and approaches. In addition, the expression ``\textit{system that can (...) make predictions, recommendations, or decisions influencing real (...) environments}'' misses out on generative AI. The updates introduced in the OECD 2023 definitions made the novel definition compliant with this premise, as the definition now reads ``\textit{system that (...) infers (...) how to generate outputs such as predictions, content, recommendations, or decisions that can influence physical or virtual environments}''.

\subsection{Scoring outcomes and recommended actions}
\label{Sec:scoring}

%TABLE%
%TABLE%
\begin{table*}
  \caption{VADER framework score outcomes, possible premise violations, and scenario-based recommended action.}
  \label{tab:vader}
  \scalebox{0.8}{
  \begin{tabular}{p{0.9in}p{1.4in}p{2.2in}p{2.3in}}
    \toprule
    Score&Premise violation&Scenario&Recommended action\\
    \midrule
    Critical risk&1. Technology neutrality&Definition includes technology&Remove references to technology\\
    % Critical risk&1. Technology neutrality&Definition would affect the market unevenly&Remove references to technology\\
    \cmidrule{2-4}
    &4. Exclusion of non-ICT&List of sufficient techniques and approaches&Remove non-exclusively-AI techniques\\
    \cmidrule{2-4}
    &5. Exclusion of non-AI ICT&No list of AI techniques and approaches& Include a list of sufficient techniques and\linebreak approaches that are exclusive to AI\\
    \midrule
    % Revision needed&1. Technology agnosticity&Definition includes technology&Remove references to technology\\
    Revision needed&2. Technical unambiguity&Definition is technically ambiguous&Adopt terminology from the literature\\
    \cmidrule{2-4}
    % &3. Technical unambiguity&Definition is technically ambiguous&Revise definition to adopt terminology from the literature\\
    % \cmidrule{2-4}
    &3. Objectivity&Definition uses subjective terms&Adopt a rational behavior perspective\\
    \cmidrule{2-4}
    &6. Inclusion of all AI&List of sufficient techniques and approaches&Include missing AI techniques and approaches\\
    &&No list of AI techniques and approaches&Include a list of sufficient techniques and\linebreak approaches that are exclusive to AI\\
    \midrule
    Appropriate \linebreak for regulation&No premises are violated&No issues can be identified&No actions needed\\
  \bottomrule
\end{tabular}
}
\end{table*}

To assist in AI policy-making, our proposed framework presents three possible outcomes that indicate to regulators how to address their proposed AI definitions~(Table \ref{tab:vader}). Below, we detail each outcome that require action, mapping how they relate to the premises previously discussed and listing recommendations for regulators.

\begin{description}[style=unboxed,leftmargin=0px,topsep=5pt]
\item[Critical risk.] The definition poses critical risk for the AI market, non-AI ICT systems, and/or works from non-ICT fields. Three premise violations result in this outcome. The first is technology neutrality, a successful principle from other technology regulation. In this case, the definition should be revised to remove references to technology in ways that would affect the market unevenly. The remaining two premise violations concern scope. If the proposed definition includes a (non)-exhaustive list of AI techniques and approaches to qualify a system as AI, the list should be revised to remove the techniques that are not exclusive to AI systems. Conversely, if no such list is being adopted, regulators should include one, provided it only comprises techniques and approaches that are exclusive to AI.

\item[Revision needed.] The definition does not pose critical risk, but should be revised to better meet its objectives. Three premise violations result in this outcome. The first regards technology unambiguity, a principle observed in other successful technology regulation. 
% In the first case, the recommended action for regulators is to remove references to technologies from the definition. In the second case, 
In this case, the definition should be revised to only include terminology that can be referenced from appropriate technical literature. The other two premise violations that result in this outcome concern scope, namely that the definition is not objective and/or that it does not include all AI techniques and approaches. In the first case, the recommended action is to revise the definition to adopt a rational behavior perspective approach. In the second case, if the proposed definition includes a (non)-exhaustive list of AI techniques and approaches to qualify a system as AI, the list should be revised to include techniques and approaches that are absent. If no list is adopted, regulators should include one, provided it only comprises techniques and approaches that are exclusive to AI.

% \item[Appropriate for regulation.] The proposed definition complies with all the seven premises comprising the VADER framework. Therefore, the definition is considered appropriate and no further action is required from regulators.
\end{description}

Table~\ref{tab:evaluation} evaluates the OECD and the remaining AI policy definition proposals as a function of the premises detailed above, where premises are given as rows and different policy (recommendations) as columns, grouped by proponent organization or country. The first three columns map score outcomes to potential premise violations. The last row depicts the VADER score for each definition. The remaining policy proposal definitions will be introduced and discussed in Section~\ref{Sec:critical}. As previously discussed, the OECD 2019 definition violates premises ``\textit{4.~Exclusion of non-ICT works}'', ``\textit{5.~Exclusion of non-AI ICT techniques, approaches, and systems}'', and ``\textit{6.~Inclusion of all AI techniques and approaches}''. In turn, the 2023 definition can be considered an improvement over the 2019 definition, since it introduces compliance to premise ``\textit{6.~Inclusion of all AI techniques and approaches}''. Nonetheless, the VADER score for both proposals is that they present \textbf{critical risk} of overregulation, even with the previously discussed improvement to premise compliance introduced by the 2023 updates. Regarding recommended actions, both definitions fit the scenario of not having a (non)-exhaustive list of AI techniques and approaches to qualify a system as AI. In this context, the easiest solution for policymakers is to include such a list, provided it only comprises techniques and approaches that are exclusive to AI.

\newcommand{\yes}{\ding{51}}
\newcommand{\no}{\ding{55}}

%TABLE%
\begin{table*}
  \caption{Evaluation of policy (recommendations), given as columns, according to the premises discussed in this work, given as rows. The first three columns map score outcomes to potential premise violations. The last row depicts the VADER score for each proposal. CR: critical risk; RN: revision needed, and; AR: appropriate for regulation.}
  \label{tab:evaluation}
  \scalebox{0.8}{
  \begin{tabular}{p{1.5in}|p{0.2in}p{0.2in}p{0.2in}|p{0.2in}p{0.2in}|p{0.24in}|p{0.32in}|p{0.32in}|p{0.38in}}
    \toprule
    &\multicolumn{3}{c|}{VADER}&\multicolumn{2}{c|}{OECD}&US&EU&UK&Brazil\\
    \midrule
    &CR&RN&AR&2019&2023&AAA&AI Act&AI Bill&PL 2338\\
    \midrule
    % 1. Technology agnosticity&\ding{51}&\ding{51}&\ding{51}& \ding{51}&\ding{51}&\ding{51}&&---&\ding{51}\\
    1. Technology neutrality&\no&\yes&\yes&\yes&\yes&\yes& \yes&\yes&\yes\\
    2. Technology unambiguity&&\no&\yes&\yes&\yes&\yes& \yes&\yes&\no\\
    \midrule
    3. Objectivity&&\no&\yes&\yes&\yes&\yes&\yes&\no&\yes\\
    4. Exclusion of non-ICT&\no&\yes&\yes&\no&\no&\no&\yes&\no&\no\\
    5. Exclusion of non-AI ICT&\no&\yes&\yes&\no&\no&\no&\yes&\no&\no\\
    6. Inclusion of all AI&&\no&\yes&\no&\yes&\no&\no&\yes&\yes\\
    \midrule
    \multicolumn{4}{r|}{\textbf{VADER score}} & \textbf{CR} & \textbf{CR} & \textbf{CR} & \textbf{RN} & \textbf{CR} & \textbf{CR} \\
  \bottomrule
\end{tabular}
}
\end{table*}

As discussed above, defining AI in an appropriate way for regulation is challenging. In this section, we have proposed a framework to assist policymakers in this task. Our proposal scores definitions based on their compliance to premises that should underlie an appropriate definition of AI for regulation, which we identify from the technical and policy-making literature. For each score outcome, our framework details the premises that may be in violation, and proposes recommended actions for regulators as a function of the scenario observed. Furthermore, our framework includes a dataset of representative AI, non-AI ICT, and non-ICT examples that help regulators address issues in scope. The case study of OECD definitions demonstrates that the application of our framework is straightforward and that the resulting benefits heavily outweigh the risks of implementing policies using the current definitions. In the next section, we use our proposed framework to critically review the AI proposal definitions from key Western and Global South AI players.

\section{A critical review of AI definitions in key regulation proposals}
\label{Sec:critical}

In the wake of the successful worldwide data privacy regulation efforts~\cite{GDPR2016a,us2018ccpa,br2018lgpd} and amid the breakthrough results displayed by deep learning algorithms~\cite{LeCun2015}, regulators around the world shifted their attention to AI. Since the first efforts in this direction, a number of proposals have been observed from key AI players. In Western societies, major examples concern the United States~\cite{us2019aaa,us2022aaa,us2023aaa,us2023eo,us2023airia,us2024ccpa}, the United Kingdom~\cite{uk2023bill}, and the European Union~\cite{eu2021aiact}. Expectedly, countries in the Global South also proposed regulation on the topic, with Brazilian efforts~\cite{br2019pl5051,br2019pl5691,br2020pl21,br2023pl2338} leading to an already preliminarily approved bill~\cite{br2023pl2338}. Worringly, these proposals all adopt different AI definitions and their appropriateness is still to be systematically evaluated from a common framework perspective. In this section, we use our proposed VADER framework for this assessment. We start with a timeline and summary of the proposals for context, the latter provided as Appendix~\ref{Sec:Summary} for brevity. Later, we detail and assess each proposal definition.  

\subsection{Timeline}

\begin{figure}[!t]
  \centering
  \includegraphics[width=0.47\linewidth]{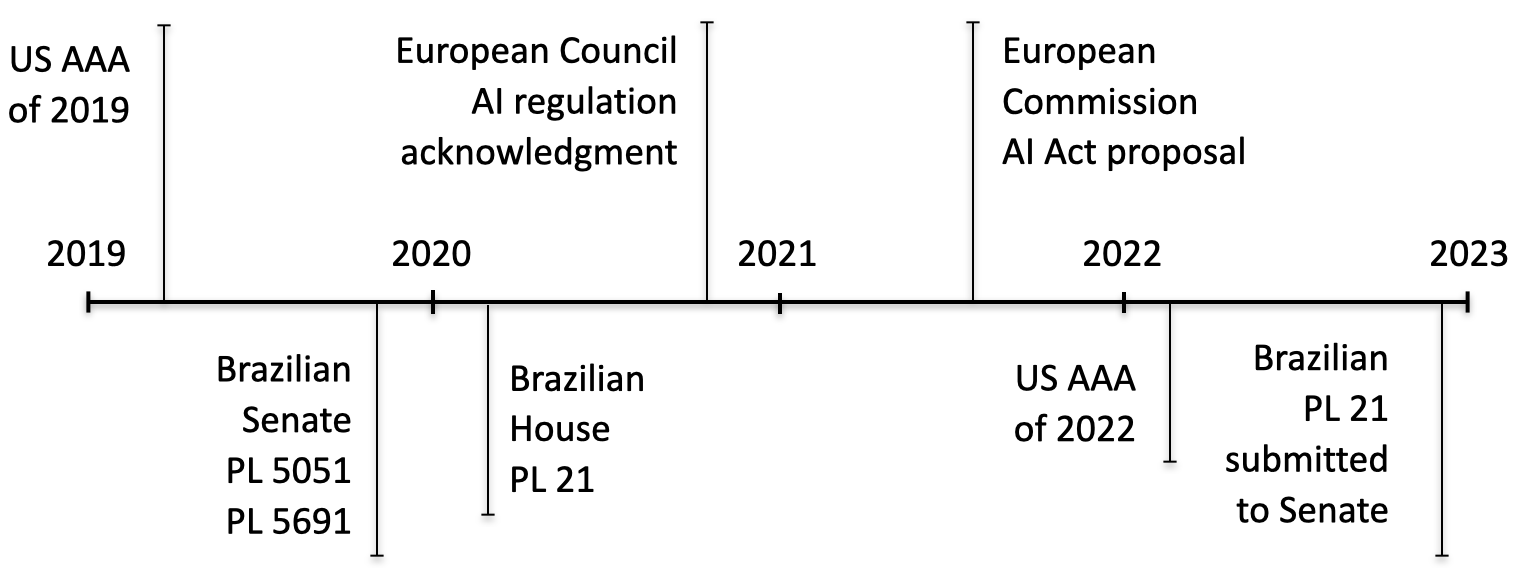}\label{fig:20192022}
  \includegraphics[width=0.52\linewidth]{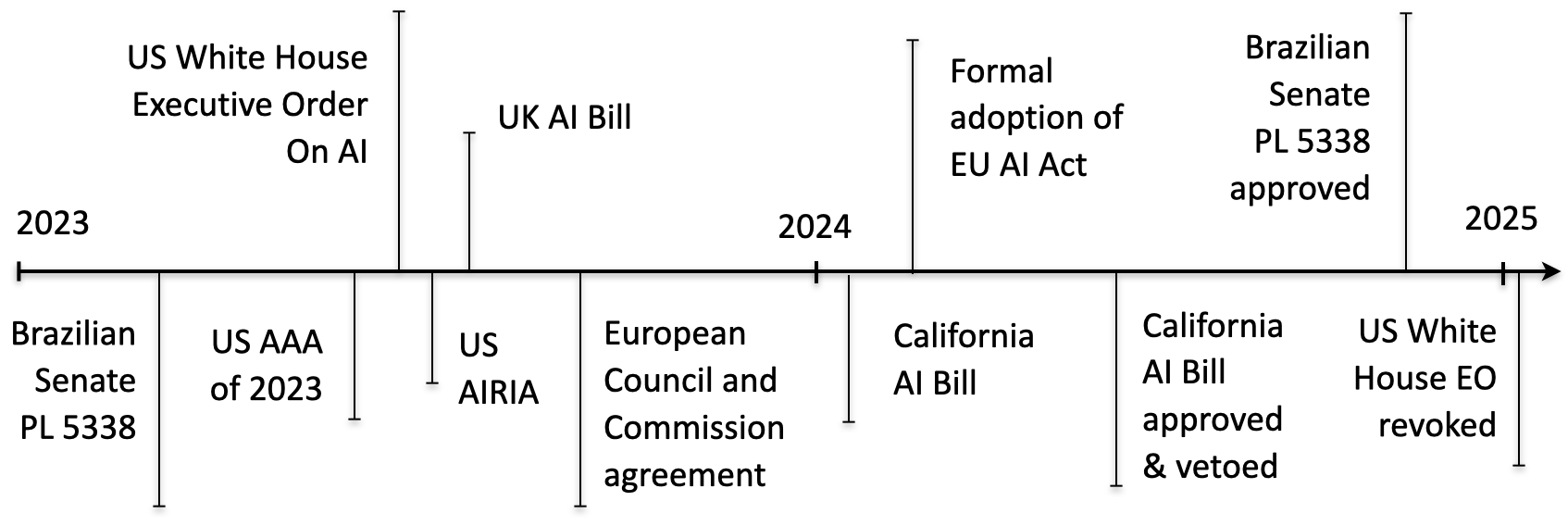}\label{fig:2023}
  \caption{Timeline of AI regulation proposals from key AI countries since 2019. Left: 2019-2022. Right: 2023 onwards.}  
  \label{fig:timeline}
\end{figure}

Figure~\ref{fig:timeline} depicts the AI timeline regulation in the countries we focus on this paper. For brevity, we remark that our scope is regulation proposals that have or may become law. As seen in Figure~\ref{fig:timeline}~(left), leading Western societies started the discussion on AI regulation back in 2019, as a natural sequence to successful data privacy regulation and in the wake of the breakthroughs enabled by deep learning~\cite{LeCun2015}. In that year, the United States had the first version of its Algorithmic Accountability Act (AAA) introduced to Congress~\cite{us2019aaa}. Similar initiatives were observed across the world around the same time, with the Brazilian Senate introducing AI Bills also in 2019~\cite{br2019pl5051,br2019pl5691}. In 2020, the Brazilian House of Representatives introduced its own AI Bill\cite{br2020pl21}, which was approved in 2022 and forwarded to the Brazilian Senate. Also in 2020, the European Council acknowledged the need for AI regulation, which led to the 2021 European Commission proposal of the EU AI Act~\cite{eu2021aiact}. In 2022, the AAA was revised and reintroduced to the US Congress~\cite{us2022aaa}, and the European Council agreed to its position on the EU AI Act.

%FIGURE%

% \begin{figure}[h]
%   \centering
%   \includegraphics[width=0.5\linewidth]{fig2}
%   \caption{Timeline of AI regulation proposals from key AI countries after May 2023.}  
%   \label{fig:2023}
% \end{figure}

The release of ChatGPT~\cite{openai2022chatgpt} at the end of 2022 and its immediate disruption to society led to a number of developments along 2023, as shown in Figure~\ref{fig:timeline}~(right). In May, the President of the Brazilian Senate introduced a proposal that strongly mirrors the EU AI Act~\cite{br2023pl2338}, which was later unified with the remaining Brazilian proposals. In September, the AAA was once again reintroduced to the US Congress, and in October the White House issued the \textit{Executive order on the Safe, Secure, and Trustworthy Development and Use of Artificial Intelligence}~(EO AI~\cite{us2023eo}) in October. The EO AI was issued a few days before the Global AI Safety Summit hosted by the UK in early November, where 30 countries signed the previously discussed Bletchley Declaration~\cite{uk2023bletchley}. Still in November, the UK government introduced its AI Bill~\cite{uk2023bill}, and the \textit{Artificial Intelligence Research, Innovation, and Accountability Act of 2023} (AIRIA~\cite{us2023airia}) was introduced to the US Congress (US, 2023c). Finally, the European Council and Parliament reached a provisional agreement on the EU AI Act in December. 

In 2024, the EU AI Act was officially adopted by the European Commission and the Brazilian Senate Bill was approved and sent to the House. In the meantime, the lack of progress in federal legislation in the US was countered by the proposal and approval of the California Safety AI Bill~\cite{us2024ccpa}. However, more recent developments in the US have been against regulation, with the California governor vetoing the state bill in September 2024 and the White House Executive Order being revoked immediately at the inauguration of the new presidential administration in January 2025.  

\subsection{Evaluation}

For the evaluation we conduct in this section, we select the four major proposals among those discussed above. In detail, we start with the (i)~the \textit{Algorithmic Accountability Act}~(AAA~\cite{us2023aaa}), the federal proposal that has been (re)introduced repeatedly in the United States. Later, we summarize (ii)~the~European Union's \textit{Artificial Intelligence Act}~(EU AI Act~\cite{eu2021aiact}) and (iii)~the~United Kingdom's \textit{Artificial Intelligence Regulation Act}~(UK AI Bill~\cite{uk2023bill}). Finally, our discussion of AI regulation in Brazil focuses on (iv)~the Brazilian Senate's PL 2338/2023~(PL2338~\cite{br2023pl2338}), as it unifies Brazilian proposals and was recently approved by the Senate.

\subsubsection*{US Algorithmic Accountability Act~\cite{us2023aaa}}
``\textit{any system, software, or process (including one derived from machine learning, statistics, or other data processing or artificial intelligence techniques and excluding passive computing infrastructure) that uses computation, the result of which serves as a basis for a decision or judgment.}'' 

\noindent\textbf{Discussion}. The AAA definition is not directly an AI definition, but rather a definition of \textit{automated decision systems}. Though this is a broad definition in itself, the enumeration of techniques and approaches in parenthesis attempts to narrow its scope. However, the enumeration does not prevent violations of premises ``\textit{4.~Exclusion of non-ICT works}'' and ``\textit{5.~Exclusion of non-AI ICT techniques, approaches, and systems}'' as, respectively, (i)~the~term ``\textit{statistics}'' includes non-ICT examples and (ii)~the~term ``\textit{data processing}'' includes non-AI ICT examples. Furthermore, since ``\textit{artificial intelligence techniques}'' are never defined with the exception of machine learning, this definition also violates premise ``\textit{6.~Inclusion of all AI techniques and approaches}''. 

\noindent\textbf{Score and recommended action}. The AAA definition is scored as \textbf{critical risk} given the violations to premises ``\textit{4.~Exclusion of non-ICT works}'' and ``\textit{5.~Exclusion of non-AI ICT techniques, approaches, and systems}''. The definition should be revised so that the enumeration used includes all and only \mbox{AI-exclusive} techniques and approaches, which would also make the definition compliant with premise ``\textit{6.~Inclusion of all AI techniques and approaches}''.

\subsubsection*{EU AI Act~\cite{eu2021aiact}}
Article~3~(1):``\textit{machine-based system that is designed to operate with varying levels of autonomy and that may exhibit adaptiveness after deployment, and that, for explicit or implicit objectives, infers, from the input it receives, how to generate outputs such as predictions, content, recommendations, or decisions that can influence physical or virtual environments''}. 
Recital~12~(excerpt): ``\textit{The techniques that enable inference while building an AI system include machine learning approaches that learn from data how to achieve certain objectives, and logic- and knowledge-based approaches that infer from encoded knowledge or symbolic representation of the task to be solved.''}

\noindent\textbf{Discussion.} The EU AI Act definition provided in its Article~3 is a rewriting of the OECD 2023 definition, differing mostly by word ordering. However, Recital~12 deepens the definition provided in Article~3, with the goal of excluding works that are non-ICT or non-AI ICT. In doing so, the EU AI Act definition becomes at the same time compliant with premises ``\textit{4.~Exclusion of non-ICT works}'' and ``\textit{5.~Exclusion of non-AI ICT techniques, approaches, and systems}'', though non-compliant with premise ``\textit{6.~Inclusion of all AI techniques and approaches}''. In detail, the excerpt from Recital~12 above serves as a list of AI techniques, but mentions only machine-learning and logic- and knowledge-based approaches.

\noindent\textbf{Score and recommended action.} The EU AI Act definition is scored as \textbf{revision needed} given the violations to premise ``\textit{6.~Inclusion of all AI techniques and approaches}''. The definition should be revised so that the enumeration used includes all and only \mbox{AI-exclusive} techniques and approaches.

\noindent\textbf{Remark.} It is important to acknowledge that the definition in the AI Act that was approved by the EU, discussed above, is an improvement over the definition adopted in the first proposal of the AI Act. In the original definition, the 2019 OECD definition was adopted, and the list of AI techniques included works from other fields such as optimization and statistics. This highlights the importance of addressing AI definition in regulation to prevent overregulation.

\subsubsection*{UK AI Bill~\cite{uk2023bill}}
``\textit{technology enabling the programming or training of a device or software to: perceive environments through the use of data; interpret data using automated processing designed to approximate cognitive abilities; and, make recommendations, predictions or decisions; with a view to achieving a specific objective. Includes generative AI, meaning deep or large language models able to generate text and other content based on the data on which they were trained.}''

\noindent\textbf{Discussion.} The UK AI Bill shares resemblance with the overall structure of the OECD 2019 definition and its corresponding premise violations. However, three major modifications are observed: (i)~replacing the term ``\textit{machine-based system}'' with ``\textit{technology enabling the programming or training of a device or software}''; (ii)~adopting the term ``cognitive abilities'', and; (iii)~including generative AI. The modifications in (i)~are~not enough to make the definition compliant with premises ``\textit{5.~Exclusion of non-ICT works}'' and ``\textit{6.~Exclusion of non-AI ICT techniques}'', as non-ICT and non-AI ICT examples are respectively still not excluded. Worryingly, the modification in (ii)~violates premise ``\textit{4.~Objectivity}'', as cognitive abilities comprise a subjective topic that stirs complex discussions. However, the inclusion of generative AI in (iii)~makes the definition compliant with premise ``\textit{7.~Inclusion of all AI techniques and approaches}''.

\noindent\textbf{Score and recommended action.} The UK AI Bill definition is scored as \textbf{critical risk} given the violations to premises ``\textit{5.~Exclusion of non-ICT works}'' and ``\textit{6.~Exclusion of non-AI ICT techniques, approaches, and systems}''. The definition should be revised to include a list of all and only \mbox{AI-exclusive} techniques and approaches. In addition, the definition also violates premise ``\textit{4.~Objectivity}''. In this context, the definition should be revised to adopt a rational behavior perspective.

\subsubsection*{Brazilian PL 2338~\cite{br2023pl2338}} AI system: ``\textit{machine-based system which, with varying levels of autonomy and for explicit or implicit objectives, infers, from either a dataset or information it is provided, how to produce outputs, particularly predictions, content, recommendations, or decisions that may influence the virtual, physical or real environment.}'' 
AI~agents:~``\textit{developers, distributors, and deployers that take part in the supply chain and in the internal governance of AI systems (...)}''

\noindent\textbf{Discussion}. The Brazilian Senate Bill definition is a rewriting of the OECD 2023 definition, differing mostly by word ordering and the definition of inputs. 
As such, the Brazilian definition violates premises ``\textit{4.~Exclusion of non-ICT works}'' and ``\textit{5.~Exclusion of non-AI ICT techniques, approaches, and systems}'',  as previously discussed for the OECD 2023 definition. Worryingly, the final version approved by the Brazilian Senate violates premise ``\textit{2.~Technical unambiguity}'', as it adopts the term AI agents to refer to people and organizations rather than to AI systems.

\noindent\textbf{Score and recommended action.} The Brazilian PL 2338 definition is scored as \textbf{critical risk} given the violation to premises ``\textit{4.~Exclusion of non-ICT works}'' and ``\textit{5.~Exclusion of non-AI ICT techniques, approaches, and systems}''. The definition should be revised to include a list of all and only \mbox{AI-exclusive} techniques and approaches. In addition, the definition also violates premise ``\textit{2.~Technical unambiguity}'', and should be revised to adopt terms from the literature.

\noindent\textbf{Remark.} The version approved by the Brazilian Senate, discussed above, violates premises different to the premises violated by the original proposal of the Bill. In the original definition, an enumeration of AI techniques was provided. Albeit non-exhaustive, the original enumeration helped prevent overregulation to non-ICT and non-AI ICT works. This highlights the importance of frameworks such as the one proposed in the work, to avoid the unintended consequence of making a novel definition compliant with some premises but non-compliant with others.

% \subsubsection*{Brazilian PL 2338~\cite{br2023pl2338} (free translation)} ``\textit{computational system, with varying levels of autonomy, designed to infer how to reach a given set of objectives, using approaches based on machine learning and/or logic or knowledge representation, from input from machines or humans, with the aim of making predictions, recommendations, or decisions that may influence the virtual or real environment.}'' 

% \noindent\textbf{Discussion}. The Brazilian PL 2338 is a combination of the OECD 2019 and 2023 definitions. The Brazilian definition improves over the definitions of OECD in that (i)~the~term ``\textit{computational system}'' replaces the term ``\textit{machine-based system}'' and that (ii)~an~enumeration of AI techniques and approaches in included. Altogether, these two modifications make the definition compliant with premises ``\textit{5.~Exclusion of non-ICT works}'' and ``\textit{6.~Exclusion of non-AI ICT techniques}''. However, the definition is not compliant with premise ``\textit{7.~Inclusion of all AI techniques and approaches}'', as it does not include examples in domains such as problem solving and search nor accounts for generative AI.

% \noindent\textbf{Score and recommended action.} The Brazilian PL 2338 definition is scored as \textbf{revision needed} given the violation to premise ``\textit{7.~Inclusion of all AI techniques and approaches}''. The definition should be revised so that the enumeration used includes all and only AI-exclusive techniques and approaches.

\section{Conclusions}
\label{Sec:Conclusion}

Artificial intelligence~(AI) is one of the technologies with most potential to disrupt social and commercial relations in the near future. Over the past decade, the breakthroughs seen from deep learning~\cite{LeCun2015} and multimodal generative AI systems like ChatGPT~\cite{openai2022chatgpt} have greatly raised the awareness, the interest, and the worries of society. As a result, AI regulation efforts that started in the wake of successful data privacy regulation gained significant momentum, and since 2023 alone at least six proposals were introduced or approved in the United States~\cite{us2023aaa,us2023eo,us2023airia}, United Kingdom~\cite{uk2023bill}, European Union~\cite{eu2021aiact}, and Brazil~\cite{br2023pl2338}. This rapid development in regulation evidences the worldwide interest in AI policymaking, ranging from Western societies to the Global South.
As important and pressing as AI regulation may be, the risk incurred by overregulation is also significant. The nature of this risk is multidisciplinary, among which (i)~\textit{legal}, from complicated norms and enforcement setup, and; (ii)~\textit{economical}, from abusive fines, costly compliance overhead, and/or obstacles to innovation. These aspects have been and will continue to be extensively debated as the regulation proposals mature worldwide. Yet, a simple and relatively overlooked risk of overregulation is scope misdefinition. In the context of AI, a multidisciplinary field with a range of academic definitions, sometimes including complex and subjective terms like intelligence and consciousness, the misconception risk cannot be overstated.

The VADER framework proposed here can be instrumental for this discussion. Comprising premises from the technical and policymaking literature, our approach enables a grounded assessment of worldwide AI regulation, scoring them according to their appropriateness for regulation as measured by premise compliance. Importantly, the framework includes a representative example dataset to help identify and address issues in scope, and provides action recommendations for policymakers to improve their proposals. 
Indeed, the assessment of the current major regulation proposals from key AI players helped both (i)~demonstrate the benefits of our proposed framework and (ii)~evidence the critical misdefinition issues with those proposals. In general, the baseline definitions that guide individual proposals come from the Organisation for Economic Co-operation and Development~(OECD). The potential issues with the original 2019 OECD proposal~\cite{oecd2019ai} were acknowledged by most countries, in that their regulation proposals include modifications to the original definition in an attempt to improve its appropriateness for regulation. Indeed, even OECD introduced an updated definition in 2023~\cite{oecd2023ai}, albeit posterior to some proposals addressed here.

The fact that \textbf{none of the six AI definition proposals evaluated in this work were scored as appropriated for regulation} as they currently stand reveals that defining scope is still an issue for regulators and policymakers worldwide. More importantly, \textbf{five of the six proposals were considered to pose concrete risk of overregulation} just because of AI misdefinition. The scope overregulation is particularly worrysome because it may affect (i)~ICT systems that in no way make use of AI, and/or; (ii)~entire fields of study other than ICT, most notably mathematics, statistics, and engineering.
Addressing the risk of overregulation by misdefinition of scope will require a collaborative effort involving governments, academia, industry, and civil society. For academia, one possible research path is to address the role played by data in regulation proposals, as clearly some of the issues presented by the AI definitions stem from trying to cover concepts such as rational agent behavior and data-driven behavior within a single definition. Another important research direction is to employ the VADER framework to a greater number of countries and to parts of the world that we have not addressed in this work. Importantly, the dataset of representative examples is a collaborative effort for which we welcome contributions, and our online, publicly available VADER repository will be available for feedback from key stakeholders and society in general.

% \section{Authors and Affiliations}

% \section{Acknowledgments}

% \begin{acks}
%  \textbf{Should we suppress for anonimity?}
% \end{acks}

%%
%% The next two lines define the bibliography style to be used, and
%% the bibliography file.
\newpage
\bibliographystyle{ACM-Reference-Format}
\bibliography{bib}

%%% -*-BibTeX-*-
%%% Do NOT edit. File created by BibTeX with style
%%% ACM-Reference-Format-Journals [18-Jan-2012].

\begin{thebibliography}{52}

%%% ====================================================================
%%% NOTE TO THE USER: you can override these defaults by providing
%%% customized versions of any of these macros before the \bibliography
%%% command.  Each of them MUST provide its own final punctuation,
%%% except for \shownote{}, \showDOI{}, and \showURL{}.  The latter two
%%% do not use final punctuation, in order to avoid confusing it with
%%% the Web address.
%%%
%%% To suppress output of a particular field, define its macro to expand
%%% to an empty string, or better, \unskip, like this:
%%%
%%% \newcommand{\showDOI}[1]{\unskip}   % LaTeX syntax
%%%
%%% \def \showDOI #1{\unskip}           % plain TeX syntax
%%%
%%% ====================================================================

\ifx \showCODEN    \undefined \def \showCODEN     #1{\unskip}     \fi
\ifx \showDOI      \undefined \def \showDOI       #1{#1}\fi
\ifx \showISBNx    \undefined \def \showISBNx     #1{\unskip}     \fi
\ifx \showISBNxiii \undefined \def \showISBNxiii  #1{\unskip}     \fi
\ifx \showISSN     \undefined \def \showISSN      #1{\unskip}     \fi
\ifx \showLCCN     \undefined \def \showLCCN      #1{\unskip}     \fi
\ifx \shownote     \undefined \def \shownote      #1{#1}          \fi
\ifx \showarticletitle \undefined \def \showarticletitle #1{#1}   \fi
\ifx \showURL      \undefined \def \showURL       {\relax}        \fi
% The following commands are used for tagged output and should be
% invisible to TeX
\providecommand\bibfield[2]{#2}
\providecommand\bibinfo[2]{#2}
\providecommand\natexlab[1]{#1}
\providecommand\showeprint[2][]{arXiv:#2}

\bibitem[AI(2024)]%
        {holistic2024state}
\bibfield{author}{\bibinfo{person}{Holistic AI}.} \bibinfo{year}{2024}\natexlab{}.
\newblock \bibinfo{booktitle}{\emph{The state of global AI regulations in 2024}}.
\newblock \bibinfo{type}{{T}echnical {R}eport}.
\newblock


\bibitem[Aranha et~al\mbox{.}(2022)]%
        {AraCamCam2022}
\bibfield{author}{\bibinfo{person}{Claus Aranha}, \bibinfo{person}{Christian~L. Camacho~Villalón}, \bibinfo{person}{Felipe Campelo}, \bibinfo{person}{Marco Dorigo}, \bibinfo{person}{Rubén Ruiz}, \bibinfo{person}{Marc Sevaux}, \bibinfo{person}{Kenneth Sörensen}, {and} \bibinfo{person}{Thomas Stützle}.} \bibinfo{year}{2022}\natexlab{}.
\newblock \showarticletitle{Metaphor-based metaheuristics, a call for action: the elephant in the room}.
\newblock \bibinfo{journal}{\emph{Swarm Intelligence}} (\bibinfo{year}{2022}), \bibinfo{pages}{1--6}.
\newblock
\urldef\tempurl%
\url{https://doi.org/10.1007/s11721-021-00202-9}
\showDOI{\tempurl}


\bibitem[Biden(2023)]%
        {us2023eo}
\bibfield{author}{\bibinfo{person}{Joseph~R Biden}.} \bibinfo{year}{2023}\natexlab{}.
\newblock \bibinfo{booktitle}{\emph{Executive order on the safe, secure, and trustworthy development and use of artificial intelligence}}.
\newblock
\urldef\tempurl%
\url{https://www.whitehouse.gov/briefing-room/presidential-actions/2023/10/30/executive-order-on-the-safe-secure-and-trustworthy-development-and-use-of-artificial-intelligence/}
\showURL{%
\tempurl}


\bibitem[Bismarck(2020)]%
        {br2020pl21}
\bibfield{author}{\bibinfo{person}{Representative~Eduardo Bismarck}.} \bibinfo{year}{2020}\natexlab{}.
\newblock \bibinfo{booktitle}{\emph{Projeto de Lei n° 21, de 2020}}.
\newblock
\urldef\tempurl%
\url{https://www.camara.leg.br/propostas-legislativas/2236340}
\showURL{%
\tempurl}


\bibitem[Brasil(2018)]%
        {br2018lgpd}
\bibfield{author}{\bibinfo{person}{Brasil}.} \bibinfo{year}{2018}\natexlab{}.
\newblock \bibinfo{booktitle}{\emph{{Lei Geral de Proteção de Dados Pessoais (LGPD)}}}.
\newblock
\urldef\tempurl%
\url{https://www.planalto.gov.br/ccivil_03/_ato2015-2018/2018/lei/l13709.htm}
\showURL{%
\tempurl}


\bibitem[Breiman(2001)]%
        {breiman2001}
\bibfield{author}{\bibinfo{person}{Leo Breiman}.} \bibinfo{year}{2001}\natexlab{}.
\newblock \showarticletitle{Statistical modeling: the two cultures}.
\newblock \bibinfo{journal}{\emph{Statist. Sci.}} \bibinfo{volume}{16}, \bibinfo{number}{3} (\bibinfo{year}{2001}), \bibinfo{pages}{199--231}.
\newblock
\showISSN{0883-4237}
\urldef\tempurl%
\url{https://doi.org/10.1214/ss/1009213726}
\showDOI{\tempurl}
\newblock
\shownote{With comments and a rejoinder by the author}.


\bibitem[{CC2020~Task~Force}(2020)]%
        {acm2020cc}
\bibfield{author}{\bibinfo{person}{{CC2020~Task~Force}}.} \bibinfo{year}{2020}\natexlab{}.
\newblock \bibinfo{booktitle}{\emph{Computing Curricula 2020: Paradigms for Global Computing Education}}.
\newblock \bibinfo{publisher}{Association for Computing Machinery}, \bibinfo{address}{New York, NY, USA}.
\newblock
\showISBNx{9781450390590}


\bibitem[Comission(2021)]%
        {eu2021aiact}
\bibfield{author}{\bibinfo{person}{European Comission}.} \bibinfo{year}{2021}\natexlab{}.
\newblock \bibinfo{booktitle}{\emph{Proposal for a Regulation of the European Parliament and of the Council Playing Down Harmonised Rules on Artificial Intelligence (Artificial Intelligence Act) and Amending Certain Union Legislative Acts}}.
\newblock
\urldef\tempurl%
\url{https://artificialintelligenceact.eu/the-act/}
\showURL{%
\tempurl}


\bibitem[Conforti et~al\mbox{.}(2014)]%
        {Conforti2014}
\bibfield{author}{\bibinfo{person}{Michele Conforti}, \bibinfo{person}{G{\'e}rard Cornu{\'e}jols}, {and} \bibinfo{person}{Giacomo Zambelli}.} \bibinfo{year}{2014}\natexlab{}.
\newblock \showarticletitle{Getting Started}.
\newblock In \bibinfo{booktitle}{\emph{Graduate Texts in Mathematics}}. \bibinfo{publisher}{Springer International Publishing}, \bibinfo{address}{Cham}, \bibinfo{pages}{1--44}.
\newblock


\bibitem[Cormen et~al\mbox{.}(2022)]%
        {Cormen2022}
\bibfield{author}{\bibinfo{person}{Thomas~H. Cormen}, \bibinfo{person}{Charles~E. Leiserson}, \bibinfo{person}{Ronald~L. Rivest}, {and} \bibinfo{person}{Clifford Stein}.} \bibinfo{year}{2022}\natexlab{}.
\newblock \bibinfo{booktitle}{\emph{Introduction to Algorithms, Fourth Edition} (\bibinfo{edition}{4th} ed.)}.
\newblock \bibinfo{publisher}{The MIT Press}.
\newblock
\showISBNx{0262033844}


\bibitem[de~Estudos~e Pesquisas Educacionais Anísio~Teixeira(2019)]%
        {inep2019cine}
\bibfield{author}{\bibinfo{person}{Instituto~Nacional de~Estudos~e Pesquisas Educacionais Anísio~Teixeira}.} \bibinfo{year}{2019}\natexlab{}.
\newblock \bibinfo{title}{Manual para classificação dos cursos de graduação e sequenciais: CINE Brasil}.
\newblock
\newblock
\urldef\tempurl%
\url{https://www.gov.br/inep/pt-br/areas-de-atuacao/pesquisas-estatisticas-e-indicadores/cine-brasil/classificacao}
\showURL{%
\tempurl}


\bibitem[Economic and Council(2023)]%
        {un2023ai}
\bibfield{author}{\bibinfo{person}{United~Nations Economic} {and} \bibinfo{person}{Social Council}.} \bibinfo{year}{2023}\natexlab{}.
\newblock \bibinfo{booktitle}{\emph{The regulatory compliance of products with embedded artificial intelligence or other digital technologies}}.
\newblock


\bibitem[{European Parliament} and {Council of the European Union}({[n.\,d.]})]%
        {GDPR2016a}
\bibfield{author}{\bibinfo{person}{{European Parliament}} {and} \bibinfo{person}{{Council of the European Union}}.} \bibinfo{year}{[n.\,d.]}\natexlab{}.
\newblock \bibinfo{booktitle}{\emph{Regulation ({EU}) 2016/679 of the {European} {Parliament} and of the {Council}}}.
\newblock
\urldef\tempurl%
\url{https://data.europa.eu/eli/reg/2016/679/oj}
\showURL{%
\tempurl}


\bibitem[for Economic Co-operation and Development(2019)]%
        {oecd2019ai}
\bibfield{author}{\bibinfo{person}{Organisation for Economic Co-operation} {and} \bibinfo{person}{Development}.} \bibinfo{year}{2019}\natexlab{}.
\newblock \bibinfo{booktitle}{\emph{Recommendation of the Council on Artificial Intelligence}}.
\newblock


\bibitem[for Statistics(2013)]%
        {unesco2013isced}
\bibfield{author}{\bibinfo{person}{UNESCO~Institute for Statistics}.} \bibinfo{year}{2013}\natexlab{}.
\newblock \bibinfo{title}{International Standard Classification of Education}.
\newblock
\newblock


\bibitem[Freund et~al\mbox{.}(2006)]%
        {Freund2006}
\bibfield{author}{\bibinfo{person}{R.J. Freund}, \bibinfo{person}{W.J. Wilson}, {and} \bibinfo{person}{P. Sa}.} \bibinfo{year}{2006}\natexlab{}.
\newblock \bibinfo{booktitle}{\emph{Regression Analysis}}.
\newblock \bibinfo{publisher}{Elsevier Science}.
\newblock
\showISBNx{9780080522975}
\urldef\tempurl%
\url{https://books.google.com.br/books?id=Us4YE8lJVYMC}
\showURL{%
\tempurl}


\bibitem[Gendreau and Potvin(2018)]%
        {Gendreau2018}
\bibfield{editor}{\bibinfo{person}{Michel Gendreau} {and} \bibinfo{person}{Jean-Yves Potvin}} (Eds.). \bibinfo{year}{2018}\natexlab{}.
\newblock \bibinfo{booktitle}{\emph{Handbook of Metaheuristics} (\bibinfo{edition}{3} ed.)}.
\newblock \bibinfo{publisher}{Springer International Publishing}, \bibinfo{address}{Cham, Switzerland}.
\newblock


\bibitem[Gill et~al\mbox{.}(2021)]%
        {GilMurWri2021}
\bibfield{author}{\bibinfo{person}{Philip~E Gill}, \bibinfo{person}{Walter Murray}, {and} \bibinfo{person}{Margaret~H Wright}.} \bibinfo{year}{2021}\natexlab{}.
\newblock \bibinfo{booktitle}{\emph{Numerical linear algebra and optimization}}.
\newblock \bibinfo{publisher}{SIAM}.
\newblock


\bibitem[Gonzalez(2009)]%
        {Gonzalez2009dip}
\bibfield{author}{\bibinfo{person}{Rafael~C Gonzalez}.} \bibinfo{year}{2009}\natexlab{}.
\newblock \bibinfo{booktitle}{\emph{Digital image processing}}.
\newblock \bibinfo{publisher}{Pearson education}.
\newblock


\bibitem[GOV.UK(2023)]%
        {uk2023bletchley}
\bibfield{author}{\bibinfo{person}{GOV.UK}.} \bibinfo{year}{2023}\natexlab{}.
\newblock \bibinfo{booktitle}{\emph{The Bletchley Declaration by Countries Attending the AI Safety Summit, 1-2 November 2023}}.
\newblock
\urldef\tempurl%
\url{https://www.gov.uk/government/publications/ai-safety-summit-2023-the-bletchley-declaration/the-bletchley-declaration-by-countries-attending-the-ai-safety-summit-1-2-november-2023}
\showURL{%
\tempurl}


\bibitem[Hoos and St{\"u}tzle(2005)]%
        {Hoos2005}
\bibfield{author}{\bibinfo{person}{H.H. Hoos} {and} \bibinfo{person}{T. St{\"u}tzle}.} \bibinfo{year}{2005}\natexlab{}.
\newblock \bibinfo{booktitle}{\emph{Stochastic Local Search: Foundations and Applications}}.
\newblock \bibinfo{publisher}{Elsevier Science}.
\newblock
\showISBNx{9781558608726}
\showLCCN{04019306}
\urldef\tempurl%
\url{https://books.google.com.br/books?id=3HAedXnC49IC}
\showURL{%
\tempurl}


\bibitem[Hyndman and Athanasopoulos(2018)]%
        {Hyndman2018}
\bibfield{author}{\bibinfo{person}{{Robin John} Hyndman} {and} \bibinfo{person}{George Athanasopoulos}.} \bibinfo{year}{2018}\natexlab{}.
\newblock \bibinfo{booktitle}{\emph{Forecasting: Principles and Practice} (\bibinfo{edition}{2nd} ed.)}.
\newblock \bibinfo{publisher}{OTexts}, \bibinfo{address}{Australia}.
\newblock


\bibitem[Inaba and Squicciarini(2017)]%
        {Inaba2017}
\bibfield{author}{\bibinfo{person}{Takashi Inaba} {and} \bibinfo{person}{Mariagrazia Squicciarini}.} \bibinfo{year}{2017}\natexlab{}.
\newblock \bibinfo{booktitle}{\emph{{ICT}: A new taxonomy based on the international patent classification}}.
\newblock \bibinfo{type}{{T}echnical {R}eport}.
\newblock


\bibitem[ISO/IEC 22989(2022)]%
        {iso2023ai}
ISO/IEC 22989 \bibinfo{year}{2022}\natexlab{}.
\newblock \bibinfo{booktitle}{\emph{Artificial intelligence concepts and terminology}}.
\newblock \bibinfo{type}{{T}echnical {R}eport}. \bibinfo{institution}{International Organization for Standardization}, \bibinfo{address}{Geneva, CH}.
\newblock


\bibitem[ISO/IEC 39794-16(2021)]%
        {iso2021ai}
ISO/IEC 39794-16 \bibinfo{year}{2021}\natexlab{}.
\newblock \bibinfo{booktitle}{\emph{Extensible biometric data interchange formats - Part 16: Full body image data}}.
\newblock \bibinfo{type}{{T}echnical {R}eport}. \bibinfo{institution}{International Organization for Standardization}, \bibinfo{address}{Geneva, CH}.
\newblock


\bibitem[Jurafsky and Martin(2018)]%
        {JurMar2018nlp}
\bibfield{author}{\bibinfo{person}{Daniel Jurafsky} {and} \bibinfo{person}{James~H Martin}.} \bibinfo{year}{2018}\natexlab{}.
\newblock \bibinfo{booktitle}{\emph{Speech and language processing}}.
\newblock
\urldef\tempurl%
\url{https://web.stanford.edu/~jurafsky/slp3/}
\showURL{%
\tempurl}


\bibitem[Lawler and Wood(1966)]%
        {Lawler1966}
\bibfield{author}{\bibinfo{person}{E.~L. Lawler} {and} \bibinfo{person}{D.~E. Wood}.} \bibinfo{year}{1966}\natexlab{}.
\newblock \showarticletitle{Branch-and-Bound Methods: A Survey}.
\newblock \bibinfo{journal}{\emph{Oper. Res.}} \bibinfo{volume}{14}, \bibinfo{number}{4} (\bibinfo{date}{aug} \bibinfo{year}{1966}), \bibinfo{pages}{699–719}.
\newblock
\showISSN{0030-364X}
\urldef\tempurl%
\url{https://doi.org/10.1287/opre.14.4.699}
\showDOI{\tempurl}


\bibitem[LeCun et~al\mbox{.}(2015)]%
        {LeCun2015}
\bibfield{author}{\bibinfo{person}{Yann LeCun}, \bibinfo{person}{Yoshua Bengio}, {and} \bibinfo{person}{Geoffrey Hinton}.} \bibinfo{year}{2015}\natexlab{}.
\newblock \showarticletitle{Deep learning}.
\newblock \bibinfo{journal}{\emph{Nature}} \bibinfo{volume}{521}, \bibinfo{number}{7553} (\bibinfo{date}{May} \bibinfo{year}{2015}), \bibinfo{pages}{436--444}.
\newblock


\bibitem[M{\"o}kander et~al\mbox{.}(2022)]%
        {Mokander2022}
\bibfield{author}{\bibinfo{person}{Jakob M{\"o}kander}, \bibinfo{person}{Prathm Juneja}, \bibinfo{person}{David~S Watson}, {and} \bibinfo{person}{Luciano Floridi}.} \bibinfo{year}{2022}\natexlab{}.
\newblock \showarticletitle{The {US} Algorithmic Accountability Act of 2022 vs. The {EU} Artificial Intelligence Act: what can they learn from each other?}
\newblock \bibinfo{journal}{\emph{Minds Mach. (Dordr.)}} \bibinfo{volume}{32}, \bibinfo{number}{4} (\bibinfo{date}{Dec.} \bibinfo{year}{2022}), \bibinfo{pages}{751--758}.
\newblock


\bibitem[Nelder and Mead(1965)]%
        {Nelder1965}
\bibfield{author}{\bibinfo{person}{J.~A. Nelder} {and} \bibinfo{person}{R. Mead}.} \bibinfo{year}{1965}\natexlab{}.
\newblock \showarticletitle{{A Simplex Method for Function Minimization}}.
\newblock \bibinfo{journal}{\emph{Comput. J.}} \bibinfo{volume}{7}, \bibinfo{number}{4} (\bibinfo{date}{01} \bibinfo{year}{1965}), \bibinfo{pages}{308--313}.
\newblock
\showISSN{0010-4620}
\urldef\tempurl%
\url{https://doi.org/10.1093/comjnl/7.4.308}
\showDOI{\tempurl}
\showeprint{https://academic.oup.com/comjnl/article-pdf/7/4/308/1013182/7-4-308.pdf}


\bibitem[of~California(2018)]%
        {us2018ccpa}
\bibfield{author}{\bibinfo{person}{The~State of California}.} \bibinfo{year}{2018}\natexlab{}.
\newblock \bibinfo{booktitle}{\emph{{California Consumer Privacy Act (CCPA)}}}.
\newblock
\urldef\tempurl%
\url{https://leginfo.legislature.ca.gov/faces/codes_displayText.xhtml?division=3.&part=4.&lawCode=CIV&title=1.81.5}
\showURL{%
\tempurl}


\bibitem[of~California Senator~Wiener(2024)]%
        {us2024ccpa}
\bibfield{author}{\bibinfo{person}{Office of California Senator~Wiener}.} \bibinfo{year}{2024}\natexlab{}.
\newblock \bibinfo{booktitle}{\emph{SB-1047 - California Legislature (2023-2024): Safe and Secure Innovation for Frontier Artificial Intelligence Models Act}}.
\newblock
\urldef\tempurl%
\url{https://leginfo.legislature.ca.gov/faces/billNavClient.xhtml?bill_id=202320240SB1047}
\showURL{%
\tempurl}


\bibitem[of~Richmond(2023)]%
        {uk2023bill}
\bibfield{author}{\bibinfo{person}{Lord~Holmes of Richmond}.} \bibinfo{year}{2023}\natexlab{}.
\newblock \bibinfo{booktitle}{\emph{Artificial Intelligence (Regulation) Bill [HL]}}.
\newblock
\urldef\tempurl%
\url{https://bills.parliament.uk/bills/3519}
\showURL{%
\tempurl}


\bibitem[of~U.S. Representative Yvette D.~Clarke(2019)]%
        {us2019aaa}
\bibfield{author}{\bibinfo{person}{Office of U.S. Representative Yvette D.~Clarke}.} \bibinfo{year}{2019}\natexlab{}.
\newblock \bibinfo{booktitle}{\emph{H.R.2231 - 116th Congress (2019-2020): Algorithmic Accountability Act of 2019}}.
\newblock
\urldef\tempurl%
\url{https://www.congress.gov/bill/116th-congress/house-bill/2231}
\showURL{%
\tempurl}


\bibitem[of~U.S. Representative Yvette D.~Clarke(2022)]%
        {us2022aaa}
\bibfield{author}{\bibinfo{person}{Office of U.S. Representative Yvette D.~Clarke}.} \bibinfo{year}{2022}\natexlab{}.
\newblock \bibinfo{booktitle}{\emph{H.R.6580 - 117th Congress (2021-2022): Algorithmic Accountability Act of 2022}}.
\newblock
\urldef\tempurl%
\url{https://www.congress.gov/bill/117th-congress/house-bill/6580}
\showURL{%
\tempurl}


\bibitem[of~U.S. Senator John~Thune(2023)]%
        {us2023airia}
\bibfield{author}{\bibinfo{person}{Office of U.S. Senator John~Thune}.} \bibinfo{year}{2023}\natexlab{}.
\newblock \bibinfo{booktitle}{\emph{S.3312 - 118th Congress (2023-2024): Algorithmic Accountability Act of 2023}}.
\newblock
\urldef\tempurl%
\url{https://www.congress.gov/bill/118th-congress/senate-bill/3312}
\showURL{%
\tempurl}


\bibitem[of~U.S. Senator Ron~Wyden(2023)]%
        {us2023aaa}
\bibfield{author}{\bibinfo{person}{Office of U.S. Senator Ron~Wyden}.} \bibinfo{year}{2023}\natexlab{}.
\newblock \bibinfo{booktitle}{\emph{S.2892 - 118th Congress (2023-2024): Algorithmic Accountability Act of 2023}}.
\newblock
\urldef\tempurl%
\url{https://www.congress.gov/bill/118th-congress/senate-bill/2892}
\showURL{%
\tempurl}


\bibitem[Office(2021)]%
        {uk2021ico}
\bibfield{author}{\bibinfo{person}{United Kingdom Information~Commisioner's Office}.} \bibinfo{year}{2021}\natexlab{}.
\newblock \bibinfo{booktitle}{\emph{Overview – Data Protection and the EU}}.
\newblock
\urldef\tempurl%
\url{https://ico.org.uk/for-organisations/data-protection-and-the-eu/overview-data-protection-and-the-eu/}
\showURL{%
\tempurl}


\bibitem[on~Artificial~Intelligence(2018)]%
        {eu2018ai}
\bibfield{author}{\bibinfo{person}{High-Level Expert~Group on Artificial~Intelligence}.} \bibinfo{year}{2018}\natexlab{}.
\newblock \bibinfo{booktitle}{\emph{A definition of AI: main capabilities and scientific disciplines}}.
\newblock \bibinfo{type}{{T}echnical {R}eport}. \bibinfo{institution}{European Commission}, \bibinfo{address}{Brussels, BE}.
\newblock


\bibitem[on~Computing~Curricula(2023)]%
        {acm2023cs}
\bibfield{author}{\bibinfo{person}{The~Joint~Task~Force on Computing~Curricula}.} \bibinfo{year}{2023}\natexlab{}.
\newblock \bibinfo{booktitle}{\emph{Computer Science Curricula 2023 Version Gamma}}.
\newblock \bibinfo{publisher}{Association for Computing Machinery}, \bibinfo{address}{New York, NY, USA}.
\newblock


\bibitem[OpenAI(2022)]%
        {openai2022chatgpt}
\bibfield{author}{\bibinfo{person}{OpenAI}.} \bibinfo{year}{2022}\natexlab{}.
\newblock \bibinfo{booktitle}{\emph{Chat{GPT}}}.
\newblock
\urldef\tempurl%
\url{https://chat.openai.com}
\showURL{%
\tempurl}


\bibitem[Pacheco(2023)]%
        {br2023pl2338}
\bibfield{author}{\bibinfo{person}{Senator~Rodrigo Pacheco}.} \bibinfo{year}{2023}\natexlab{}.
\newblock \bibinfo{booktitle}{\emph{Projeto de Lei n° 2338, de 2023}}.
\newblock
\urldef\tempurl%
\url{https://www25.senado.leg.br/web/atividade/materias/-/materia/157233}
\showURL{%
\tempurl}


\bibitem[Russell et~al\mbox{.}(2023)]%
        {oecd2023ai}
\bibfield{author}{\bibinfo{person}{Stuart Russell}, \bibinfo{person}{Karine Perset}, {and} \bibinfo{person}{Marko Grobelnik}.} \bibinfo{year}{2023}\natexlab{}.
\newblock \bibinfo{booktitle}{\emph{Updates to the OECD’s definition of an AI system explained}}.
\newblock
\urldef\tempurl%
\url{https://oecd.ai/en/wonk/ai-system-definition-update}
\showURL{%
\tempurl}


\bibitem[Russell and Norvig(2020)]%
        {russell2020}
\bibfield{author}{\bibinfo{person}{Stuart~J. Russell} {and} \bibinfo{person}{Peter Norvig}.} \bibinfo{year}{2020}\natexlab{}.
\newblock \bibinfo{booktitle}{\emph{Artificial Intelligence: a modern approach} (\bibinfo{edition}{4} ed.)}.
\newblock \bibinfo{publisher}{Pearson}.
\newblock


\bibitem[Sampat and Zavala(2019)]%
        {Sampat2019}
\bibfield{author}{\bibinfo{person}{Apoorva~M Sampat} {and} \bibinfo{person}{Victor~M Zavala}.} \bibinfo{year}{2019}\natexlab{}.
\newblock \showarticletitle{Fairness measures for decision-making and conflict resolution}.
\newblock \bibinfo{journal}{\emph{Optim. Eng.}} \bibinfo{volume}{20}, \bibinfo{number}{4} (\bibinfo{date}{Dec.} \bibinfo{year}{2019}), \bibinfo{pages}{1249--1272}.
\newblock


\bibitem[Turing(1950)]%
        {Turing1950}
\bibfield{author}{\bibinfo{person}{Alan~M. Turing}.} \bibinfo{year}{1950}\natexlab{}.
\newblock \showarticletitle{Computing Machinery and Intelligence}.
\newblock \bibinfo{journal}{\emph{Mind}} \bibinfo{volume}{59}, \bibinfo{number}{236} (\bibinfo{year}{1950}), \bibinfo{pages}{433--460}.
\newblock
\showISSN{00264423, 14602113}
\urldef\tempurl%
\url{http://www.jstor.org/stable/2251299}
\showURL{%
\tempurl}


\bibitem[Valentim(2019a)]%
        {br2019pl5051}
\bibfield{author}{\bibinfo{person}{Senator~Styvenson Valentim}.} \bibinfo{year}{2019}\natexlab{a}.
\newblock \bibinfo{booktitle}{\emph{Projeto de Lei n° 5051, de 2019}}.
\newblock
\urldef\tempurl%
\url{https://www25.senado.leg.br/web/atividade/materias/-/materia/138790}
\showURL{%
\tempurl}


\bibitem[Valentim(2019b)]%
        {br2019pl5691}
\bibfield{author}{\bibinfo{person}{Senator~Styvenson Valentim}.} \bibinfo{year}{2019}\natexlab{b}.
\newblock \bibinfo{booktitle}{\emph{Projeto de Lei n° 5691, de 2019}}.
\newblock
\urldef\tempurl%
\url{https://www25.senado.leg.br/web/atividade/materias/-/materia/139586}
\showURL{%
\tempurl}


\bibitem[Wolpert and Macready(1997)]%
        {WolMac1997opt}
\bibfield{author}{\bibinfo{person}{D.H. Wolpert} {and} \bibinfo{person}{W.G. Macready}.} \bibinfo{year}{1997}\natexlab{}.
\newblock \showarticletitle{No free lunch theorems for optimization}.
\newblock \bibinfo{journal}{\emph{IEEE Transactions on Evolutionary Computation}} \bibinfo{volume}{1}, \bibinfo{number}{1} (\bibinfo{year}{1997}), \bibinfo{pages}{67--82}.
\newblock
\urldef\tempurl%
\url{https://doi.org/10.1109/4235.585893}
\showDOI{\tempurl}


\bibitem[Wolpert(1996)]%
        {Wol1996learning}
\bibfield{author}{\bibinfo{person}{David~H. Wolpert}.} \bibinfo{year}{1996}\natexlab{}.
\newblock \showarticletitle{The lack of a priori distinctions between learning algorithms}.
\newblock \bibinfo{journal}{\emph{Neural Computation}} \bibinfo{volume}{8}, \bibinfo{number}{7} (\bibinfo{year}{1996}), \bibinfo{pages}{1341--1390}.
\newblock
\urldef\tempurl%
\url{https://doi.org/10.1162/neco.1996.8.7.1341}
\showDOI{\tempurl}


\bibitem[Wolpert(2001)]%
        {Wol2001sl}
\bibfield{author}{\bibinfo{person}{David~H. Wolpert}.} \bibinfo{year}{2001}\natexlab{}.
\newblock \showarticletitle{The supervised learning no-free-lunch theorems}. In \bibinfo{booktitle}{\emph{Proceedings of the 6th Online World Conference on Soft Computing in Industrial Applications}}.
\newblock


\bibitem[Wolpert and Macready(1995)]%
        {WolMac1995search}
\bibfield{author}{\bibinfo{person}{David~H. Wolpert} {and} \bibinfo{person}{William~G. Macready}.} \bibinfo{year}{1995}\natexlab{}.
\newblock \bibinfo{booktitle}{\emph{No free lunch theorems for search}}.
\newblock \bibinfo{type}{{T}echnical {R}eport} SFI-TR-95-02-010. \bibinfo{institution}{Santa Fe Institute}, \bibinfo{address}{Sante Fe, NM, USA}.
\newblock


\end{thebibliography}

%%
%% If your work has an appendix, this is the place to put it.
\newpage
\appendix

\section{Systems and systems-of-systems taxonomies}
\label{Sec:SoS}

The plurality in ICT systems is further evidenced by the huge number of taxonomies that has been published since~2000, which provide systematic ways to understand and categorize concepts in this broad and dynamic field. One of the most recent ICT taxonomies is the ``J tag''~\cite{Inaba2017}, based on the technology classes of the International Patent Classification~(IPC). The ``J tag'' taxonomy includes concepts from other existing taxonomies, such as the \textit{Organisation for Economic Co-operation and Development}~(OECD) ICT~taxonomy~(2003), and their revised definitions of the ICT~sector~(2007) and of ICT~products~(2008). Concretely, the ``J tag'' taxonomy defines 13 technology areas and related ICT~products, as depicted in Table~\ref{tab:Jtag}. The proposed taxonomy is broad and covers all technology areas addressing systems~(hardware and software components). In complement to ``J tag'', other system taxonomy proposals can be referred to as well, but most of them covers a specific aspect of ICT.
%, such as an ICT~taxonomy for network, an ICT~taxonomy for cybersecurity%
%~\cite{Fovino2019}
%, an ICT~taxonomy for information systems, and others. 

%TABLE%
\begin{table*}
  \caption{The "J tag" taxonomy of ICT systems \cite{Inaba2017}}
  \label{tab:Jtag}
  \scalebox{0.8}{
  \begin{tabular}{p{2in}p{5in}}
    \toprule
    \textbf{Technology areas}&\textbf{Products}\\
    \midrule
    High speed network & Digital transmission, network (protocols, architecture, etc.), telephone communication, broadcasting, and transmission, reception, channels\\    
\midrule
    Mobile communication&Cellular systems, wireless Local Area Networks (LAN) and Personal Area Networks (PAN). \\
\midrule
    Security&Secret-coding, authentication, and electronic payment  \\
\midrule
    Sensor and device network& Ubiquitous Sensor Networks\\
\midrule
    High speed computing& Computer architecture, composition of hardware (arithmetic, logic, control, input/output, and storage units), computer programs, and operating systems\\
\midrule
    Large-capacity and high speed storage&Various storage device-related technologies (e.g. semiconductor memory, magnetic storage, optical storage, etc.); network (e.g. network attached storage, NAS; storage area network, SAN); and file systems\\
\midrule
    Large-capacity information analysis& Database and numerical analysis, computational science, and computer aided engineering \\
\midrule
    Cognition and meaning understanding&Cognitive computing\\
\midrule
    Human-interface&Human-interface technologies\\
\midrule
    Imaging and sound technology&Video equipment, television, image processing, acoustic equipment, and audio signal processing-related technologies \\
\midrule
    Information communication device&Electronic circuits, communication cables, semiconductor lasers, etc. \\
\midrule
    Electronic measurement&Radio navigation, radio direction-finding, etc\\    
  \bottomrule
\end{tabular}
}
\end{table*}

As seen in Table~\ref{tab:Jtag}, the range of technology areas comprised by ICT systems poses a significant challenge when appropriately defining AI for regulation. Even more so, current real-world systems fit the definition of systems~of~systems~(SoS), in that they comprise a set of constituent independent and interconnected systems to achieve goals that independently none would be able to achieve.  These types of systems are commonly found across diverse domains and industries, including telecommunications, financial, healthcare, transportation, defense, infrastructure, and others. ISO/IEC/IEEE~21841:2019 is a well-known standardized SoS taxonomy based on their degree of managerial and operational independence. Four SoS taxa are defined: (i)~directed, (ii)~acknowledged, (iii)~collaborative, and (iv)~virtual. A~\textit{directed}~SoS has a central decision-making authority, whereas an~\textit{acknowledged}~SoS has a shared purpose and goal, with the individual constituent systems maintaining independent control and objectives. In~\textit{collaborative}~SoS, the constituent systems operate without the obligation to adhere to centralized management, choosing instead to participate voluntarily in a collaborative effort to achieve the common goal. Finally, \textit{virtual}~SoS are those that operate without centralized managerial control or a shared common purpose. 
These diverse taxa pose further additional challenges for accountability and enforcement in the context of AI regulation, which we plan to address in future work.
%In the context of AI regulation, these diverse taxa pose further additional challenges for accountability and enforcement. However, we leave this discussion for future work, as it exceeds the scope of the current paper. 
\section{Taxonomy of AI techniques and approaches}
\label{Sec:AI}

As discussed in the background review, the \textit{Association for Computing Machinery}~(ACM) classifies artificial intelligence~(AI) as a computer science~(CS) discipline. The \textit{ACM Computer Science Curriculum 2023}~\cite{acm2023cs} follows closely the seminal work of~\citet{russell2020}, which explores a set of concepts that would underlie the development of an intelligent agent (or system). Very briefly, an intelligent agent that is able to perceive the world and act upon it would comprise the following properties:

\begin{description}[style=unboxed,leftmargin=0px]
    \item[Problem solving and search.] Given a predetermined goal and a set of inputs, an intelligent agent should be able to define a sequence of actions that would maximise a performance metric. This process is called search. Search algorithms can be further classified into uninformed and informed~(heuristics). The former, also known as blind search, fits the remit of the \textit{algorithms and complexity} CS discipline~\cite{Cormen2022}, does not exploit specific knowledge about the problem domain, and explores the search space until the goal state is found. Examples include depth-first search and breadth-first search. In the latter, the search algorithm exploits additional information about the problem or an heuristic function, a function that estimates the cost~(distance) to reach the goal state from a given problem state. Here, there is a group of algorithms that, instead of searching all the possible paths to the best solution, focus on (stochastic) local search strategies that disregard the path to the goal and iteratively probe the environment, thus incrementally moving towards a satisfactory solution - not necessarily optimal~\cite{Hoos2005}. These algorithms excel in large or complex problems, in which exhaustive search strategies are unfeasible. Examples include swarm intelligence and evolutionary algorithms~\cite{Gendreau2018}.

    \item[Knowledge representation, reasoning and planning.] To solve complex problems or reach good solutions, a desirable property of an intelligent agent is the ability to form representations about the world, based on a representation language, its syntax and semantics. Knowledge further enables reasoning, and novel representations of the world can be inferred, supported by diverse logic paradigms. Finally, the intelligent agent can establish~(plan) the sequence of actions that maximizes its goal. Examples of methods include rule-based systems and expert systems, decision trees, and graph-based models. This process may happen in uncertain, nondeterministic, or partially observable scenarios, in which case the agent could rely on probabilistic reasoning. Methods include Bayesian networks, hidden Markov models, and Kalman filters. In this context, techniques employed are deeply connected to philosophy and logic, whether classical or fuzzy. 
    
    \item[Learning.] A fundamental property of intelligent agents is the ability to improve their performance on a task based on its prior experiences and observations about the world. Learning takes place in different forms, the most usual of which are: supervised learning, in which the input-action pair is available~(labeled data); unsupervised learning, in which the system finds patterns or relationships in the unlabeled data; and reinforcement learning, in which learning is based on a reward or punishment signal following the agent’s action. Examples for (i)~supervised, (ii)~unsupervised, and (iii)~reinforcement learning include, respectively: (i)~k-nearest neighbours, support vector machines, and random forests; (ii)~k-means clustering, principal component analysis, and autoencoders, and; (iii)~Q-learning and Monte Carlo search. Notably, neural networks are a relevant example of learning technique that can be applied for all three types of learning. In addition, agents can refer to techniques originating in statistics when the properties of the data distribution are understood, such as linear and logistic regression, and the (S)AR(I)MA(X) family of algorithms~\cite{breiman2001}. 

    \item[Natural language.] To harness human knowledge an intelligent agent needs to cope with the human language in which that knowledge is expressed. The human language can be encoded as written text or as speech databases, and stirs multiple tasks, such as text classification, information retrieval, and question answering. From a high-level perspective, natural language processing~(NLP) analytical approaches address language from a lexical or structural perspective, using techniques such as tokenization and parsing. In complement, natural language understanding~(NLU) approaches deal with semantics, using techniques such as language models. 
    
    \item[Perception.] An intelligent agent is expected to interact with its environment and other agents. For that, it is fundamental to be able to sense the world, passively or actively, extracting information to solve problems, search, reason, plan, and learn. In other words, perceiving the world is the link between the environment and the agent’s internal processes. Examples include speech, vision, and diverse sensory information. In terms of techniques, analytical and mathematical approaches are adopted in image and signal processing algorithms, whereas understanding involves techniques such as visual embeddings in computer vision.
    
    \item[Robotics.] Intelligent agents need not operate solely in a virtual world. Robots equipped with sensors and actuators are capable of interacting with the physical world. The original view of robots as manufacturing entities devoted to repetitive tasks in controlled environments has been replaced by untethered autonomous agents with sophisticated sensors and actuators that can work alone or as part of a swarm, on the ground, in the air, or underwater. In fact, there is a whole group of AI researchers, in particular those within the embodied cognitive science field, that reinforce that truly intelligent agents could only be developed by taking into account brain-body-environment interactions, thus posing robotics as a central topic in AI development. Examples range from classic control algorithms to brain-inspired simultaneous localization and mapping~(SLAM) strategies.

\end{description}

As highlighted above, AI techniques and approaches span over a broad range of topics. More importantly, for each such topic, techniques and approaches employed may not be exclusive to AI systems, as the field is inherently multidisciplinary. Altogether, these observations motivate the example dataset we propose in this paper, which help differentiate between \textit{sufficient} and \textit{non-AI-exclusive} examples of techniques and approaches used in AI systems.
%TABLE%
\begin{table*}
  \caption{Mapping between existing AI policy recommendations and AI perspectives. Policy recommendations are sorted from top to bottom from the most recent to the oldest.}
  \label{tab:mapping}
  \begin{tabular}{rcp{0.5in}p{0.5in}p{0.5in}p{0.5in}}
    \toprule
    Policy recommendation & Year & Human behavior & Human thought & Rational thought & Rational behavior\\
    \midrule
    OECD~\cite{oecd2023ai} & 2023  &--&--&--& \ding{51}\\
    The Bletchley Declaration~\cite{uk2023bletchley} & 2023 &--&--&--& --\\
    UN Economic and Social Council~\cite{un2023ai} & 2023 &--&--&--& \ding{51}\\
    ISO/IEC EN 22989~\cite{iso2023ai} & 2022 &--&--&--& \ding{51}\\
    ISO/IEC 39794-16~\cite{iso2021ai} & 2021 &\ding{51}&--&--&--\\
    OECD~\cite{oecd2019ai} & 2019 &--&--&--& \ding{51}\\
    EU High-Level Expert Group on AI~\cite{eu2018ai} & 2018 &--&--&\ding{51}&\ding{51}\\
  \bottomrule
\end{tabular}
\end{table*}

\section{AI policy recommendations and definition perspectives}
\label{Sec:Definitions}

Table \ref{tab:mapping} maps existing AI policy recommendations to the perspectives discussed in Section~\ref{Sec:Background}. Policy recommendations are sorted from top to bottom from the most recent to the oldest. Notice that the Bletchley declaration does not include an AI definition, despite being one of the most recent among all assessed~\cite{uk2023bletchley}. In addition, the EU High-Level Expert Group on AI definition can be mapped to multiple perspectives~\cite{eu2018ai}. In general, however, the rational behavior perspective is the one most commonly employed. Importantly, even the same organization can propose different definitions of AI, as observed for the International Standards Organization~(ISO). In detail, ISO/IEC EN 22989 primarily addresses AI and adopts a rational behavior perspective~\cite{iso2023ai}. Conversely, ISO/IEC 39794-16:2021 adopts a human behavior perspective when addressing extensible biometric data interchange formats~\cite{iso2021ai}. Finally, we remark that the AI definition originally proposed by OECD in 2019~\cite{oecd2019ai} has already been revised in 2023~\cite{oecd2023ai} to account for novel developments in the field. Importantly, this has led to a discrepancy between definitions adopted by OECD and~(aspiring to be) member countries.
%, as we will later discuss.

\section{Data privacy regulations}
\label{Sec:Regulation}

In this section, we detail two successful examples of technological regulation, which build on the same principles previously discussed for GDPR~\cite{GDPR2016a}: (i)~the~California Consumer Privacy Act~(CCPA\cite{us2018ccpa}), and; (ii)~the~Brazilian General Data Protection Law (LGPD~\cite{br2018lgpd}).

\begin{description}[style=unboxed,leftmargin=0px]
    \item[CCPA.] In the United States, there is a notable absence of comprehensive data protection legislation~\cite{holistic2024state}. Instead, a patchwork of federal and state laws addresses specific aspects of data privacy, creating a fragmented regulatory landscape. The collected data of many companies remains largely unregulated across states, with federal laws typically applicable only to government entities and specific sectors.%
    \footnote{\url{https://www.forbes.com/sites/conormurray/2023/04/21/us-data-privacy-protection-laws-a-comprehensive-guide/?sh=22a341e85f92 }}
    State laws vary significantly, introducing legal uncertainty, especially for businesses operating nationally. Efforts to establish federal data privacy legislation span fifty years~\cite{Mokander2022}, prompting government administrations to resort to executive orders and other legal instruments in the absence of a comprehensive framework. In this context, the California Consumer Privacy Act~(CCPA) stands out as a pivotal data privacy law in the United States. Despite lacking a nationwide scope, the law holds global implications, influencing international discussions on privacy regulations.%
    \footnote{\url{https://www.economist.com/united-states/2019/02/28/congress-is-trying-to-create-a-federal-privacy-law}} 
	
    Regarded as the most stringent US data privacy law, the CCPA intentionally mirrors the GDPR. In detail, the CCPA represents the first US law aiming to provide transparency in the use of personal data and grants consumers control over their personal information. In addition, the CCPA is crafted to be technologically neutral, intended to be applicable to all emerging technologies without showing preference for any specific methods or tools, thus aligning with the evolving landscape of ICT. The legislature's intent is to apply the law to various industries operating within California and globally, and the tech neutrality further facilitates this objective. Transparency, consumer empowerment in personal data control, and the promotion of data security are key tenets of the CCPA, reflecting its commitment to fostering a privacy-centric environment in line with contemporary technological advancements.

    \item[LGPD.] In Brazil, the General Data Protection Law~(LGPD) has drawn inspiration from established regulations, notably the GDPR and the CCPA, resulting in significant similarity to both frameworks. LGPD also has an extraterritorial applicability, as it extends its reach to any operation involving personal data collected within Brazilian territory or belonging to a Brazilian citizen. This means that it is not only applied to entities or individuals who processed data inside Brazil, but also to those outside the country that handle personal data related to a Brazilian resident. By incorporating such provisions, the LGPD positions Brazil within the broader context of evolving global norms and reinforces its commitment to safeguarding the privacy and data rights of its citizens.
	
    Aligned with the codes previously highlighted, the LGPD incorporates \textit{privacy by design} as a foundational principle, ensuring transparent data processing activities that respects the rights of data subjects. To oversee compliance, the law establishes the National Data Protection Authority~(ANPD) and outlines obligations for companies, which are subject to the entity. The code employs language and broad definitions that aim to be technologically neutral, avoiding favoring or excluding specific types of technology. The law emphasizes fundamental data protection principles rather than prescribing detailed technical measures, allowing the organizations and individuals to develop and implement technologies and practices aligned with these principles. It is important to note that the LGPD also provides a legal framework adaptable to technological advances acknowledging the dynamic nature of ICT.

\end{description}

\section{Dataset of representative examples for scope validation}
\label{Sec:Dataset}

Tables \ref{tab:Non-ICT}--\ref{tab:AI} respectively list non-ICT, non-AI ICT, and sufficient AI examples.

\begin{table*}[h]
  \caption{Non-ICT examples.}
  \label{tab:Non-ICT}
  \begin{tabular}{p{2in}p{1.5in}p{2in}}
    \toprule
   System or Process&Domain&Reference\\
    \midrule
    Simplex Algorithm&Mathematical optimization&\citet{Nelder1965}\\
    Integer Linear Programming&Mathematical optimization&\citet{Conforti2014}\\
    Linear Regression&Statistics&\citet{Freund2006}\\
    Logistic Regression&Statistics&\citet{Freund2006}\\
    (S)AR(I)MA(X) Models&Statistics&\citet{Hyndman2018}\\
    Thermostat & Electronic Engineering & \citet{russell2020}\\
    Steam Engine Governor&Mechanical Engineering&\citet{russell2020}\\
    Water clock with a regulator that \linebreak maintained  a constant flow rate&Mechanical Engineering&\citet{russell2020}\\
    \bottomrule
\end{tabular}
\end{table*}

\begin{table*}[h]
  \caption{Examples of ICT techniques, approaches, and systems that, in themselves, are not regarded as AI.}
  \label{tab:Non-AI}
  \begin{tabular}{p{2in}p{2.2in}p{1.3in}}
    \toprule
   System or Process&Domain&Reference\\
    \midrule
    Uninformed search strategies
    %: breadth-first, depth-first, etc.
    &Algorithms and Complexity&\citet{Cormen2022}\\
    Branch-and-bound&Algorithms and Complexity&\citet{Lawler1966}\\
    Enterprise resource planning&Software Engineering&\citet{acm2020cc}\\
    Web Services&Parallel and Distributed Computing&\citet{acm2020cc}\\
    Operating Systems&Operating Systems&\citet{acm2020cc}\\
    Database Management Systems & Data Management & \citet{acm2020cc}\\
    Compilers & Foundations of Programming Languages & \citet{acm2020cc}\\
    Dashboards & Information Systems & \citet{acm2020cc}\\
    \midrule
    Stemming & Natural Language Processing & \citet{JurMar2018nlp}\\
    Tokenization & Natural Language Processing & \citet{JurMar2018nlp}\\
    Lemmatization & Natural Language Processing & \citet{JurMar2018nlp}\\
    Parsing & Natural Language Processing & \citet{JurMar2018nlp}\\
    $n$-grams & Natural Language Processing & \citet{JurMar2018nlp}\\
    TF-IDF & Natural Language Processing & \citet{JurMar2018nlp}\\
    Bag-of-Words & Natural Language Processing & \citet{JurMar2018nlp}\\    
    \midrule
    Image compression & Digital Image Processing & \citet{Gonzalez2009dip}\\ 
    Mathematical filtering & Digital Image Processing & \citet{Gonzalez2009dip}\\
    Affine transformations & Digital Image Processing & \citet{Gonzalez2009dip}\\
    Mathematical morphology & Digital Image Processing & \citet{Gonzalez2009dip}\\
    Bag-of-Visual-Words & Digital Image Processing & \citet{Gonzalez2009dip}\\
        \bottomrule
\end{tabular}
\end{table*}

\begin{table*}[h]
  \caption{Sufficient AI examples of techniques and approaches.}
  \label{tab:AI}
  \begin{tabular}{p{1.5in}p{2.7in}p{1.3in}}
    \toprule
   System or Process&Domain&Reference\\
    \midrule
    Local search&Problem solving and search&\citet{Hoos2005}\\
    Stochastic local search&Problem solving and search&\citet{Hoos2005}\\
    Swarm intelligence&Problem solving and search&\citet{Gendreau2018}\\
    Evolutionary algorithms&Problem solving and search&\citet{Gendreau2018}\\
    \midrule
    Logical agents&Knowledge representation, reasoning, and planning &\citet{russell2020}\\
    Bayesian networks&Knowledge representation, reasoning, and planning &\citet{russell2020}\\
    Probabilistic reasoning&Knowledge representation, reasoning, and planning &\citet{russell2020}\\
    \midrule
    Supervised learning&Learning &\citet{russell2020}\\
    Unsupervised learning&Learning &\citet{russell2020}\\
    Reinforcement learning&Learning &\citet{russell2020}\\
    \midrule
    Textual embeddings&Natural language understanding &\citet{russell2020}\\
    Languages models&Natural language understanding &\citet{russell2020}\\
    Visual embeddings&Computer vision &\citet{russell2020}\\
    \bottomrule
\end{tabular}
\end{table*}

\section{Summary of the current AI regulation scenario}
\label{Sec:Summary}

In this section, we summarize the main characteristics of the AI regulation proposals assessed in this work. We start with the 
%(i)~\textit{Executive Order on AI}~(EO AI~\cite{us2023eo}) and 
(i)~the \textit{Algorithmic Accountability Act}~(AAA~\cite{us2023aaa}), the major federal proposal in the United States. Later, we summarize the European Union's \textit{Artificial Intelligence Act}~(EU AI Act~\cite{eu2021aiact}), the United Kingdom's \textit{Artificial Intelligence Regulation Act}~(UK AI Bill~\cite{uk2023bill}), and the Brazilian Senate's PL 2338/2023~(PL2338~\cite{br2023pl2338}).

\textbf{US AAA:} The Algorithmic Accountability Act of 2023 is a legislative proposal introduced in the Senate with the aim of safeguarding individuals subject to algorithmic decision-making in critical domains like credit, education, health, and more. 
\textit{“The Algorithmic Accountability Act of 2023 requires companies to assess the impacts of the AI systems they use and sell, creates new transparency about when and how such systems are used, and empowers consumers to make informed choices when they interact with AI systems.” } 
This legislation assigns the Federal Trade Commission (FTC) the responsibility of formulating guidelines to direct companies operating AI systems in conducting concrete assessments to mitigate social, ethical, and legal risks \cite{Mokander2022}.
Proposed in 2019, 2022, and 2023, the recurring introduction of the bill underscores the US commitment to establishing conditions for the responsible use and development of AI systems and algorithms. The discussions over effective AI regulation in the US are growing, with some experts arguing that the best way to achieve a complete regulation would be issuing several specific legislation, such as the AAA. 
Importantly, the AAA applies to new generative AI systems, automated decision systems (ADS), and other systems involved in augmented critical decision-making processes. It encompasses organizations under FTC jurisdiction across all 50 states and any US territory, reflecting a concerted effort to address the evolving challenges posed by AI technologies.

\textbf{EU AI Act:} The European Union (EU) has taken a pioneering step proposing the AI Act, a comprehensive regulation for AI technology. Introduced for the first time in 2021, the AI Act reflects a fundamental aspect of UE policy, emphasizing the promotion of secure and legal use of AI systems while respecting the fundamental rights of the citizens. The proposal follows a risk management approach, aiming to establish a horizontal and integrated legal structure for AI systems with a primary focus on legal security.
In 2022, the European board reached an initial agreement on a general framework and started cooperation with the European Parliament in early 2023. By the end of that year, the EU Act achieved the final stages of legislative procedure, with the Parliament and the Council reaching a provisional agreement. The AI Act was officially adopted in May 2024, and published in the Official Journal of the European Union a couple months later.
Structurally, the AI Act introduces a European governance and compliance system, granting autonomy to the 27 participant countries to regulate AI internally. Embracing a risk-based approach, the law categorizes AI systems into four risk levels, each subject to specific rules. The code outlines prohibited AI practices and delimitates the AI systems exempt from its purview, such as those exclusively used for military or national security purposes. Moreover, the agreement predicts that the rules therein are not applied to AI systems used solely for research, innovation, or nonprofessional purposes.
The Act establishes a horizontal safety layer to ensure that AI systems do not violate any constitutional rights or significantly impact distinct areas of the society. While prioritizing safety, the provisions also anticipate support measures for the secure development, innovation, and validation of new AI systems, striking a balance between regulation and fostering technological advancement.

\textbf{UK AI Bill (AI Regulation Act, UK):}
The UK AI Act, introduced in the UK House of Lords in the end of 2023, represents a significant stride in the United Kingdom's commitment since signing the Bletchley Declaration, earlier that year.
The code represents a general framework for AI regulation within the UK, featuring the creation of an AI Authority tasked with overseeing the path to AI regulation and formulating guidelines and principles, such as transparency and explainability, safety, robustness, fairness, and contestability, which will conduct the Authority procedure.
The bill addresses the governance of AI technology at a commercial level, introducing the role of AI responsible officers for companies involved in the development, deployment, or use of AI systems. Furthermore, the AI Authority should cooperate with various sectors to establish sandboxes, fostering an environment that encourages AI innovation and development.
The legislation also assigns specific responsibilities for the Secretary of State, who will play a pivotal role in delivering more detailed AI regulations through statutory instruments.
While the UK AI Act can be viewed as a response to the commitments made during the 2023 AI Summit, it is acknowledged that further refinement may be necessary. Nevertheless, this act marks a crucial step toward AI regulation in the UK, contributing to the global discourse on responsible AI development and deployment.

\textbf{Brazilian PL 2338.} The Brazilian legislative initiative, Bill PL 2338, is a general regulation designed to ensure that the development and utilization of AI systems are guided by ethical principles and constitutional rights. Positioned as a safeguard for users, the bill outlines their rights when interacting with AI systems and provides a framework for categorizing AI based on the potential risks they pose to society. Emphasizing the centrality of human rights, particularly privacy and free speech, the law holds AI developers and providers liable for any damages caused by their systems.
The bill sets a few key principles to guide AI development and usage, including transparency, nondiscrimination, accountability, data privacy and protection, and diversity. It promotes interoperability among AI systems, promoting the creation of ecosystems where diverse AI solutions can collaborate, facilitating integration, and advancing technology in areas such as data analysis, decision-making, and various sectors of the society.
PL 2338 mandates that all AI systems offer traceability of decision-making processes, ensuring that information is clear and accessible to all affected individuals, allowing them to comprehend and contest the decisions.
Governance is a focal point of the project, encompassing measures related to data organization for training, testing, and result validation, as well as information security throughout the AI system's lifecycle. The bill addresses transparency, bias mitigation, and dataset representativity, with additional governance measures specified for high-risk AI systems to reduce potential risks and standardize algorithmic impact assessment procedures. This legislative framework reflects a comprehensive approach to fostering responsible and ethical AI development and usage in Brazil, and has been approved by the Senate in December 2024. The Bill is now expected to be discussed by the Brazilian House of Representatives in 2025 before being sanctioned by the President.

\end{document}